%% file: main.tex
\documentclass{article}

\PassOptionsToPackage{numbers, sort&compress}{natbib}

\usepackage[final]{neurips_2024}

\newif\ifincludecontent

\usepackage[utf8]{inputenc} %
\usepackage[T1]{fontenc}    %
\usepackage{xcolor}         %
\definecolor{citecolor}{HTML}{2779af}
\definecolor{linkcolor}{HTML}{c0392b}
\usepackage[pagebackref=false,breaklinks=true,colorlinks,bookmarks=false,citecolor=citecolor,linkcolor=linkcolor]{hyperref}
\usepackage{url}            %
\usepackage{booktabs}       %
\usepackage{amsfonts}       %
\usepackage{nicefrac}       %
\usepackage{microtype}      %

\usepackage{amsmath}
\usepackage{amssymb}
\usepackage{mathtools}
\usepackage{amsthm}
\usepackage{enumitem}

\usepackage[capitalize]{cleveref}

\usepackage{todonotes}
\usepackage{xspace}
\usepackage{colortbl}
\usepackage{subcaption}
\usepackage{bbm}
\usepackage{pifont}
\usepackage{wrapfig}

\definecolor{lightred}{rgb}{1, 0.6, 0.6}
\definecolor{lightgreen}{rgb}{0.4, 0.8, 0.4}
\definecolor{mediumdarkred}{rgb}{0.8, 0.0, 0.0}

\newcommand{\greencheck}{{\color{green}\ding{51}}}
\newcommand{\redx}{{\color{red}\ding{55}}}

\newcommand{\asz}[1]{}
\newcommand{\rdh}[1]{}
\newcommand{\plan}[1]{}
\newcommand{\bogdan}[1]{}
\newcommand{\harsh}[1]{}
\newcommand{\zkn}[1]{}

\newcolumntype{P}[1]{>{\centering\arraybackslash}m{#1}}

\definecolor{lightgray}{gray}{0.95}
\definecolor{lightblue}{RGB}{180,255,180}
\definecolor{Envdisc}{RGB}{100, 149, 237} %
\definecolor{Envcont}{RGB}{205, 92, 92} %
\definecolor{lightgray}{gray}{0.95}

\newcommand{\vlms}{MLLMs\xspace}

\newcommand{\asalong}{Action Space Adapter\xspace}
\newcommand{\asa}{ASA\xspace}
\newcommand{\asas}{ASAs\xspace}
\newcommand{\rvq}{RVQ\xspace}
\newcommand{\vq}{VQ\xspace}
\newcommand{\mlp}{Pred\xspace}
\newcommand{\unif}{Uniform\xspace}
\newcommand{\semlang}{SemLang\xspace}
\newcommand{\lang}{Lang\xspace}

\newcommand{\taskcount}{114\xspace}

\newcommand{\llama}{LLaMA\xspace}
\newcommand{\llava}{LLaVA\xspace}
\newcommand{\vlm}{MLLM\xspace}
\newcommand{\rt}{RT-Inspired\xspace}
\newcommand{\scratch}{Scratch\xspace}

\newcommand{\calvin}{CALVIN\xspace}
\newcommand{\metaworld}{Meta-World\xspace}
\newcommand{\habpick}{HabPick\xspace}
\newcommand{\Habpick}{Habitat Pick\xspace}
\newcommand{\langR}{Language Rearrangement\xspace}
\newcommand{\LangR}{Language Rearrangement\xspace}
\DeclarePairedDelimiter\norm{\lVert}{\rVert}

\title{Grounding Multimodal Large Language Models \\ in Actions}

\author{
    Andrew Szot$^{1,2}$ \quad
    Bogdan Mazoure$^1$ \quad
    Harsh Agrawal$^1$ \quad
    Devon Hjelm$^{1,3}$ \\
    \textbf{Zsolt Kira$^2$} \quad
    \textbf{Alexander Toshev$^1$} \quad \\
    $^1$ Apple, $^2$ Georgia Tech, $^3$ Mila \\
    \textit{a.szot@apple.com, toshev@apple.com}
}

\begin{document}

\maketitle

\input{sections/abstract}
\input{sections/intro}

\input{sections/related_work}
\input{sections/method}

\input{sections/experiments}
\input{sections/conclusion}

{\small
\bibliographystyle{unsrtnat}
\setlength{\bibsep}{0pt}
\bibliography{main}
}

\newpage
\appendix
\input{supp/prior_work}
\input{supp/exps}

\end{document}

%% file: sections/abstract.tex
\begin{abstract}
\looseness=-1 Multimodal Large Language Models (MLLMs) have demonstrated a wide range of capabilities across many domains, including Embodied AI. In this work, we study how to best ground a MLLM into different embodiments and their associated action spaces, with the goal of leveraging the multimodal world knowledge of the MLLM. We first generalize a number of methods through a unified architecture and the lens of action space adaptors. For continuous actions, we show that a learned tokenization allows for sufficient modeling precision, yielding the best performance on downstream tasks. For discrete actions, we demonstrate that semantically aligning these actions with the native output token space of the MLLM leads to the strongest performance. We arrive at these lessons via a thorough study of seven action space adapters on five different environments, encompassing over 114 embodied tasks.
\end{abstract}

%% file: sections/intro.tex
\section{Introduction}

Multimodal Large Language Models (\vlms), defined as Large Foundation Models that take as input text and images and generate text, have recently seen rapid progress and impressive performance~\citep{bai2023qwen,kosmos-1,peng2023kosmos2,blip-2,instruct-blip,llava,li2023multimodal,zhu2023minigpt,ye2023mplug,li2023otter,li2023mimic, mckinzie2024mm1,idefics}. These models are important as they solve a large range of useful yet difficult natural language and image tasks, such as describing images, answering visual and textual questions, reasoning, and learning from a small number of examples. They have only recently improved to the point of being usable enough for general deployment with human non-experts~\citep{team2023gemini,achiam2023gpt,touvron2023llama}. 

While \vlms are capable of describing real-world embodied concepts, their capabilities in embodied tasks are limited to using text for actions through generating code~\cite{liang2022code,zeng2022socratic}, representing actions as text~\cite{brohan2023rt}, or extracting actions from internal representations~\cite{li2023vision,szot2023large}.
\emph{Grounding}~\citep{tellex2020robots} \vlms to generate actions extends their capabilities to embodied tasks, such as robot manipulation and navigation, and is of tremendous value for practical problems, potentially overcoming the high cost of training tabula rasa.
Extending \vlms to multimodal image generation enables object detection and segmentation, and image and video generation~\cite{peng2023kosmos2,chen2023shikra,you2023ferret,wang2023visionllm,lai2023lisa,zhang2023llava}.
In embodied settings, grounding \vlms via predicting agent affordances and generating actions yields effective policies capable of generalizing to new tasks~\cite{ahn2022can, szot2023large, driess2023palm, brohan2023rt}.

A key and open challenge in grounding \vlms, which limits their capabilities in embodied tasks, is the gap between the native output space, natural language, and the action space of embodied agents.
This problem is particularly acute in continuous action spaces, where low-level controllers may require a high degree of precision. Across the literature, a number of architectures and ways of handling action spaces have been proposed, but there has not been a systematic study of these designs.
Our contributions generalize prior attempts to adapt \vlms to generate actions through an empirical study on which principles and strategies are necessary to effectively close the gap between the action spaces of \vlms and embodied agents. 
We study various grounding re-parameterization strategies, which we refer to as Action Space Adapters (ASAs), across a range of embodiments, action spaces, and environments. In particular, we explore the following types of \asas: (1) \asas that directly generate actions from a new prediction policy using the \vlm hidden representations as input; (2) \asas that reuse the native token space of the \vlm to encode actions; (3) and \asas that introduce a new token space to encode the actions of the agent while adapting the \vlms to predict these new tokens.

Further, we empirically identify important principles for designing \asas. For continuous action spaces, learned tokenization with several vocabularies that residually model continuous actions gives the right modeling precision while using vocabularies of manageable sizes and, as a result, yields the best performance across all continuous control environments. 
This learned tokenization outperforms direct action prediction, indicating this approach allows the model to effectively learn a multimodal distribution over action spaces. In addition, the above tokenization strategy boosts performance when the policy is a \vlm, compared to other standard non-LLM-based policies, indicating that it manages to better tap into the knowledge of the model.

For discrete action spaces, we study \asas that better align the embodied actions with the output space of the \vlm. We demonstrate that a semantic alignment between these -- mapping discrete actions to semantically related tokens in the \vlm vocabulary -- yields the best strategy compared to other adapters that either reuse or define a new vocabulary. The superiority of this strategy is evident in performance on environments with discrete action spaces and also in RL sample efficiency.

\begin{figure*}[t!] 
  \centering
  \includegraphics[width=\textwidth]{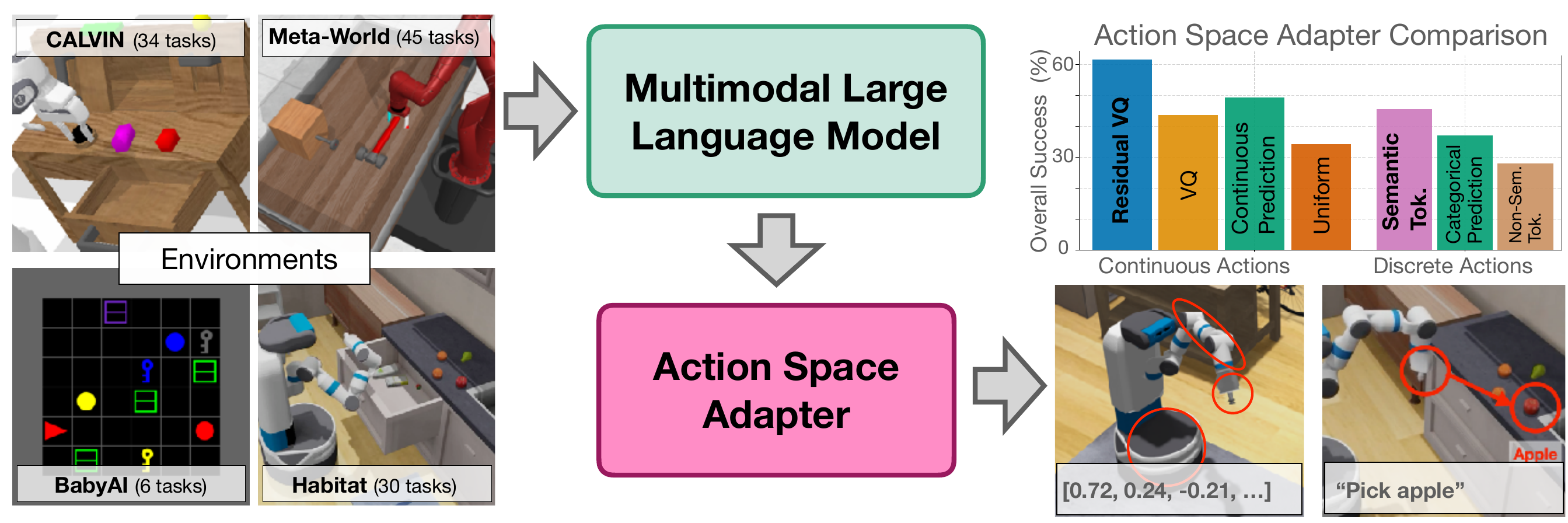}
  \caption{
    We empirically analyze how to ground \vlms in actions across \taskcount tasks in continuous and discrete action spaces. 
    In each environment, we train a multi-task policy with different {\asalong}s (\asas) to re-parameterize the \vlm to output actions. For continuous actions, learning a tokenization with several tokens per-action performs best (Residual VQ). For discrete actions, mapping actions to semantically related language tokens performs best (Semantic Tokenization).
  }
  \label{fig:teaser}
\end{figure*}

Finally, the above principles are thoroughly validated across five embodied AI environments, three of which are robotic continuous control and two with discrete actions as illustrated in \Cref{fig:teaser}.
Altogether, we consider \taskcount language specified tasks. 
In the continuous case, the best tokenization achieves $72\%$ on \calvin~\cite{mees2022calvin}, up from $68\%$ for direct action regression and $28\%$ for uniform action tokenization; and $84\%$ on \metaworld~\cite{yu2020meta}, up from $61\%$ for direct action regression and $75\%$ for uniform tokenization. Similarly, in the case of discrete actions, the proposed semantically aligned action tokens yield $51\%$ on LangR~\cite{szot2023large}, up from $42\%$ for direct action prediction.

%% file: sections/related_work.tex
\section{Related Work}
\label{sec:related-work} 

\looseness=-1 Prior works propose different {\asalong}s (\asas) to adapt \vlms into policies.
Some works use LLMs or \vlms as zero-shot policies by prompting them to output text or code that can be executed as actions~\citep{zeng2022socratic,shah2023lm,huang2022inner,liang2023code,huang2023grounded,wu2023tidybot,silver2023generalized,wang2023voyager}.
The \asa in this case is a given executor or low-level controller that takes text as input and outputs actions in the environment.
Other works investigate adapting \vlms for actions, but focus on a single \asa and environment.
For example, RT-2~\cite{brohan2023rt} uniformly discretizes continuous actions and predicts tokens corresponding to each of the action dimensions.
RoboFlamingo~\cite{li2023vision}, Lamo~\cite{shi2023unleashing}, and LLaRP~\cite{szot2023large} use an MLP to predict an environment action from an LLM hidden state.
GFlan~\cite{carta2023grounding} treats discrete actions as text and ranks actions by the LLM log probability to form a distribution over actions.
At a high level, our work is distinct in that we study a variety of methods across multiple environments for learning \asas. We focus on tasks with low zero-shot VLM performance, such as low-level control or long-horizon planning tasks. 
We summarize the differences between our investigation and prior work adapting VLMs for action in \Cref{sec:prior-work}.

Investigating action representations in embodied settings is not new. 
Some works learn representations of actions to help generalization to new actions or operating in large action spaces \cite{jain2020generalization,dulac2015deep} in the context of Reinforcement Learning (RL).
Our study proposes \asas for tokenizing continuous actions, and other works use different types of discretization or tokenization strategies on continuous action spaces.
\cite{shafiullah2022behavior,cui2022play} use k-means to discretize continuous actions to help learn from multimodal behavior datasets, such as from play data or data from different experts.
VQ-BeT~\cite{lee2024behavior} finds learning a residual VQA (RVQ) codebook for continuous actions works best but does not apply this idea to \vlms.

More broadly, prior works have adapted \vlms for modalities other than actions, such as object bounding boxes and image generation, both being continuous in nature while the latter of high dimension. For example, 
\cite{peng2023kosmos,zhang2023llava} train \vlms to output spatial reference tokens to ground text responses in image regions.
For image generation, \cite{sun2023generative} adapt \vlms to generate image patches;  
\cite{yu2023scaling,aiello2023jointly} tokenize images using a VQ-VAE model and adapt \vlms to generate images by decoding these image tokens, which has inspired us to use the same learned tokenization; 
\cite{lee2022autoregressive} uses an RVQ model ~\cite{zeghidour2021soundstream} to generate images, similarly to our best performing tokenization scheme.

%% file: sections/method.tex
\section{Method}

\begin{figure*}[t]
  \centering
  \includegraphics[width=0.95\textwidth]{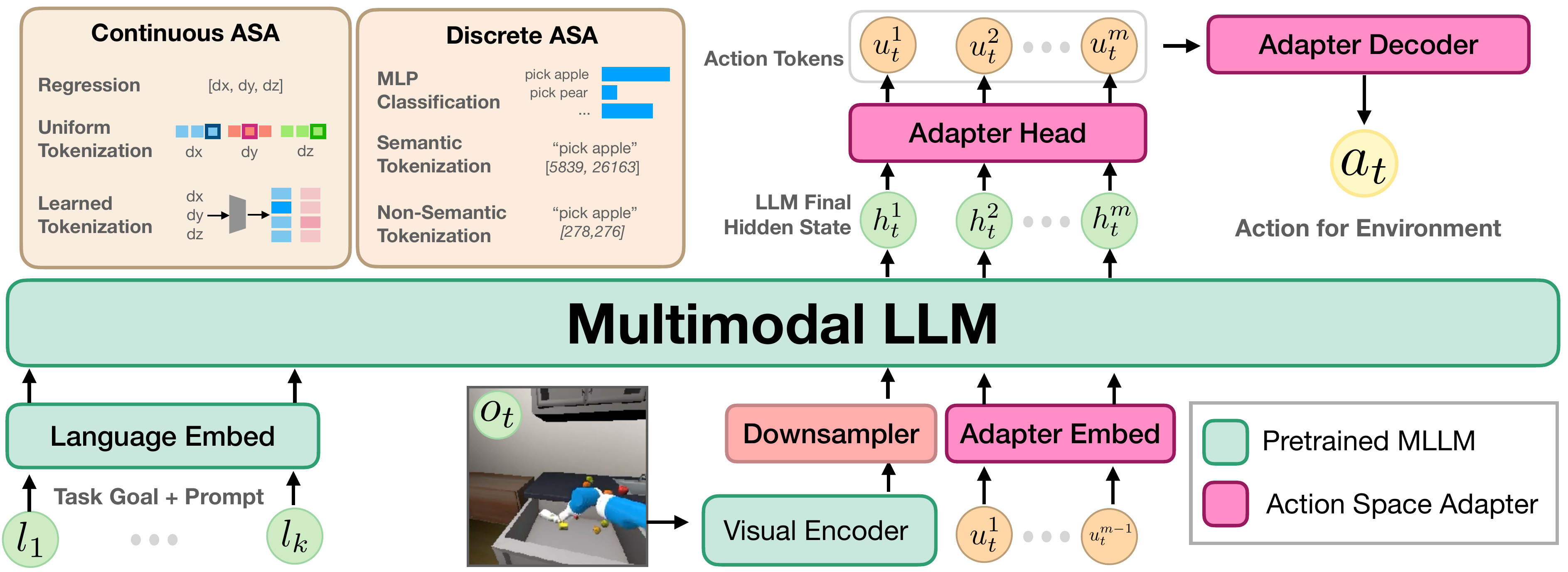}
  \caption{
    Generic architecture studied here for adapting \vlms for action-specific decision making. The \vlm takes the embedding of the task instruction, prompt, and visual tokens as input. The \vlm then autoregressively predicts a sequence of $ m$ action tokens. These action tokens are then decoded into an environment-specific action.
  }
  \label{fig:method}
\end{figure*}

In order to solve an embodied task, an agent learning in an interactive environment must select a decision from a set of valid actions. 
For example, an action space could be a set of keyboard presses for a video game or a real-valued vector that controls a robotic manipulator.
Our work studies how to best adapt a \vlm, which is originally trained to output text tokens, to instead model actions from a given environment. 
We refer to the module that bridges a \vlm with a certain action space as an \emph{\asalong} (ASA) (see \Cref{fig:method}). 

\subsection{Problem Setting}
Our analysis focuses on language-specified tasks with visual observations.
Specifically, we consider a goal-specified Partially-Observable Markov Decision Process (POMDP)~\cite{Bel} that has an observation space $ \mathcal{O}$, action space $ \mathcal{A}$,  and goal space $ \mathcal{G}$. For brevity, we omit other elements of the MDP. In our setting, $ \mathcal{G}$ is a textual description of the task to solve. $\mathcal{O}$ consists of RGB visual perception and agent proprioception. We consider a range of different action spaces $\mathcal{A}$ that broadly fall into two categories -- discrete and continuous.
The primary objective is to learn a language-conditioned policy that maps observations and the instruction text to an action $ \pi(a|o, g)$. 
As later described in \cref{sec:training}, we learn this policy through supervised fine tuning from expert demonstrations or reinforcement learning that maximizes the expected discounted cumulative reward of the POMDP.

\subsection{From Vision and Language to Action}
\label{sec:asa} 

The process studied here for adapting \vlms for decision making is illustrated in \Cref{fig:method}. 
The \vlm policy takes as input a textual instruction describing the downstream task, a sequence of past observations in the task and outputs an action in the agent's action space.
In the bottom left of \cref{fig:method}, the task description, as well as the environment description, are first encoded to produce language embeddings. 
To these embeddings, the \vlm then appends a sequence of visual embeddings from the current observation $o_t$. 
Since visual embeddings can often be comprised of a large number of tokens (the popular LLaVA-1.5 model~\cite{llava} has 556), we introduce a downsampling layer to enable the \vlm to attend over a longer history of observations. 
In practice, we take the downsampling layer to be a Perceiver model~\cite{jaegle2021perceiver}, a learnable transformation that reduces the number of tokens from the visual encoder before being used as input to the \vlm. 

The sequence of language and visual embeddings is passed through the \vlm, whose final hidden state $h_t^1$ encodes the entire input. 
The \asa, whose trainable parameters are denoted $\theta$, is comprised of three parts: (1) an adapter head, (2) an adapter embedding, and (3) an adapter decoder. 
The hidden state is first passed through the adapter head to produce action tokens $u_t^1=A_\theta(h_t^1)$. 
The action tokens are then embedded using the action embedding into $E_\theta(u_t^1)$, and passed autoregressively through the \vlm to produce further hidden embeddings $h_t^2,\ldots, u_t^m$ and associated action tokens $u_t^2, \ldots, u_t^m$, resulting in total $m$ tokens per time step. 
The predicted action tokens are then decoded into the final action $a_t$ by the adapter decoder, which produces the final action $a_t=D_\theta(u_t^1,..,u_t^m)$. 
As $a_t\in \mathcal{A}$, it is then executed in the environment to produce $o_{t+1}$, and the process continues. 

Next, we describe possible ASA implementations for discrete and continuous action spaces.

\subsubsection{Discrete Action Spaces}
\label{sec:disc-action}

We define the following action spaces adapters for a discrete action space $\mathcal{A}$:

\textbf{Categorical Prediction (\mlp)}: Implement the action space adapter as an MLP network, which predicts the logits of a categorical distribution over environment actions from the \vlm hidden state. The adapter head is an MLP that maps the hidden state $ h^{1}$ directly to an action $ a \in \mathcal{A}$. This amounts to producing a single action token $u^1$, which directly corresponds to the action $a$, with the action decoder being an identity map. Both the adapter head and token embeddings are initialized from scratch. This type of ASA is used by~\cite{szot2023large}. 

\textbf{Semantic Language (\semlang)}: The action space adapter predicts natural language text that maps to a discrete action. First, each action $ a \in \mathcal{A}$ is described with freeform text tokenized as $(l_1, \dots , l_m)$. The \vlm then autoregressively predicts a sequence of $ m$ tokens, which are then decoded by the adapter decoder to the corresponding action. For example, in an action space choosing a high-level skill $ a$ could be described as ``pick apple", which is tokenized as $ [5839, 26163] $ with the \llama tokenizer. The \vlm then must sequentially predict token $ 5839$, then token $ 26163$ to call this action. Sequences of tokens corresponding to invalid actions are either avoided entirely with the token filter described in \cref{sec:training} or treated as a no-op. Both the adapter head and the token embeddings are re-used to be the pretrained LLM's language head and embedding layer, respectively, meaning no additional parameters over the pretrained \vlm are added. This type of ASA is used by~\cite{driess2023palm}.

\textbf{Non-Semantic Language (\lang)}: Actions are mapped to language tokens, but instead of semantically meaningful descriptions of the actions as with \semlang, the actions are mapped to sequences of numbers. For example, ``pick apple" is represented with the string ``5 3". The policy must then output the tokens corresponding to this text to call this pick action. Note that we can pick any text for this mapping and the choice of integers is arbitrary. However, the selected text is not semantically representative of the action.

\subsubsection{Continuous Action Space Adaptors}
\label{sec:cont-action}
We define the following four \asas for a continuous $D$-dimensional action space $\mathcal{A}$: the first \asa predicts in the original action space while the other three use tokenization. At training time, we learn a policy to predict these action tokens from the \asa. At test time, we employ an action decoder that maps these action tokens to actions in the original space $\mathcal{A}$.

\textbf{Continuous Regression (\mlp)}: Regress to the original continuous action from the MLLM hidden state $h_t^1$. This is achieved via a single-layer MLP network, which is trained using MSE loss. This \asa is used by \cite{li2023vision,shi2023unleashing}.

\textbf{Uniform Action Tokenization (\unif)}: The simplest approach is to use uniform binning of the action space. In particular, we express each action as a sequence of $D$ tokens by quantizing each of the $D$ action dimensions into one out of $K$ uniform bins:
\[
\textrm{Uniform(a)} = (k_1\ldots k_D)\quad\textrm{such that}\quad a_d\in \textrm{bin}(k_d, d)
\]
where $\textrm{bin}(k, d)$ denotes the $k^\textrm{th}$ bin along the $d^\textrm{th}$ action dimension. If $m_d$ and $M_d$ denote the lower and upper bounds respectively of the $d^\textrm{th}$ action dimension, then its definition reads $\textrm{bin}(k, d) = [m_d + k\frac{M_d - m_d}{K}, m_d + (k+1)\frac{M_d - m_d}{K}] $. At test time, we decode predicted action tokens to the center of the corresponding bins for each dimension. This type of ASA is used by~\cite{brohan2023rt}. 

\textbf{Vector Quantized Tokenization (\vq)}: To adapt the tokenization to the particular action space, we propose to use learned tokenization. In particular, we express each action as a single token that corresponds to the closest action code from a learned codebook $V$. Using encoder network $ f_\theta$ that maps actions to a latent embedding space:
\[
\textrm{VQ}(a) = (k_1) \quad\textrm{where}\quad k_1 = \arg\min_{k}||f_\theta(a) - v_k||_2^2
\]
where $v_k\in V$. The codebook $V$ of size $K$ is learned over an offline dataset $\mathcal{D}$ of actions using a VQ-VAE~\cite{van2017neural} trained with the mean-squared error for action reconstruction and commitment loss.
We overwrite $ K$ infrequently used tokens from the LLM vocabulary to represent $ V$. We defer the full details of this tokenization process to \Cref{sec:asa-details}.

\textbf{Residual Vector Quantized Tokenization (\rvq)}: Precise control requires precise action modeling that can suffer after tokenization. To increase the precision of a learned tokenization, we further investigate the use of a sequence of several action tokens as in Uniform. Similar to VQ, these tokens are from $M$ action codebooks $V_m, m\in\{1, \ldots, M\}$. However, each codebook models the residual space obtained after modeling the action using preceding codebooks, thus each subsequent token captures increasingly finer action information:
\[
  \textrm{RVQ}(a) = (k_1, \ldots k_M)\quad\textrm{where}\quad k_m = \arg\min_k \norm*{ \left(  f_\theta(a)-\sum_{i=1}^{m-1}v^i_{k_i} \right) - v^m_k}_2^{2}
\]
where $v_k^i\in V_i$ is the $k^\textrm{th}$ code from the $i^\textrm{th}$ codebook. Such tokenization can be learned using Residual VQ-VAE~\citep[RVQ-VAE,][]{lee2022autoregressive} on an offline dataset of actions. The actual number of token sequences we can represent is $K^M$. Hence, RVQ presents the opportunity to exponentially increase the action space quantization without having to drastically increase the size of the learned individual codebooks.

\subsection{Training}
\label{sec:training} 
We use LLaVA-1.5-7B~\cite{llava} as the base \vlm. We finetune the \vlm with interactive (i.e., action-labeled) data to make it more suited for interacting with a embodied and interactive environment.

\textbf{Supervised Fine Tuning (SFT) with Expert Demonstrations:} We finetune the \vlm for interactive tasks using a dataset of expert demonstrations. Each demonstration contains (1) a language description of the task, (2) a sequence of observations, and (3) a sequence of actions that successfully solve the task. 
Note that in this work, we are primarily interested in learning imitation policies from offline data, which can be extended to offline reinforcement learning if per-timestep rewards are included in the dataset. 
Specifically, we train the \vlm with supervised learning to predict the expert actions from the observations and language description in the data. 
While the pre-trained LLM and the visual encoder remain frozen, we finetune the \asa, the visual downsampler, and parts of the LLM with LoRA~\cite{hu2021lora}. 
In total, the model has $ \approx 100M$ learnable LLM parameters and $ \approx 40M$ learnable downsampler and \asa parameters. The learned tokenization schemes (\rvq and \vq) have an additional pre-training phase, where the VAE models are first trained on actions from the offline dataset and then frozen to prevent further updates in later stages.

\textbf{Reinforcement Learning (RL) from Environment Feedback} We can also optionally finetune the \vlm to optimize an environment reward using RL. However, predicting actions in the \vlm token space dramatically increases the number of possible action predictions, with many possible predictions corresponding to no valid action. For example, there are 32,000 tokens in the \llama text tokenizer, giving $ 32,000^{m}$ possible predictions by the model with $ m$ tokens per action. This makes exploration difficult in RL as only a small fraction of the possible actions are valid. We therefore use a \emph{token filter} to restrict the autoregressive sampling to only be from token sequences corresponding to valid actions. The token filter is a function $ M(l_t^{1}, \dots, l_t^{j-1}) $ that produces a binary mask over all tokens to represent valid tokens for the $ j$th decoding step.

%% file: sections/experiments.tex
\begin{figure*}[t!] 
  \centering
  \includegraphics[width=\textwidth]{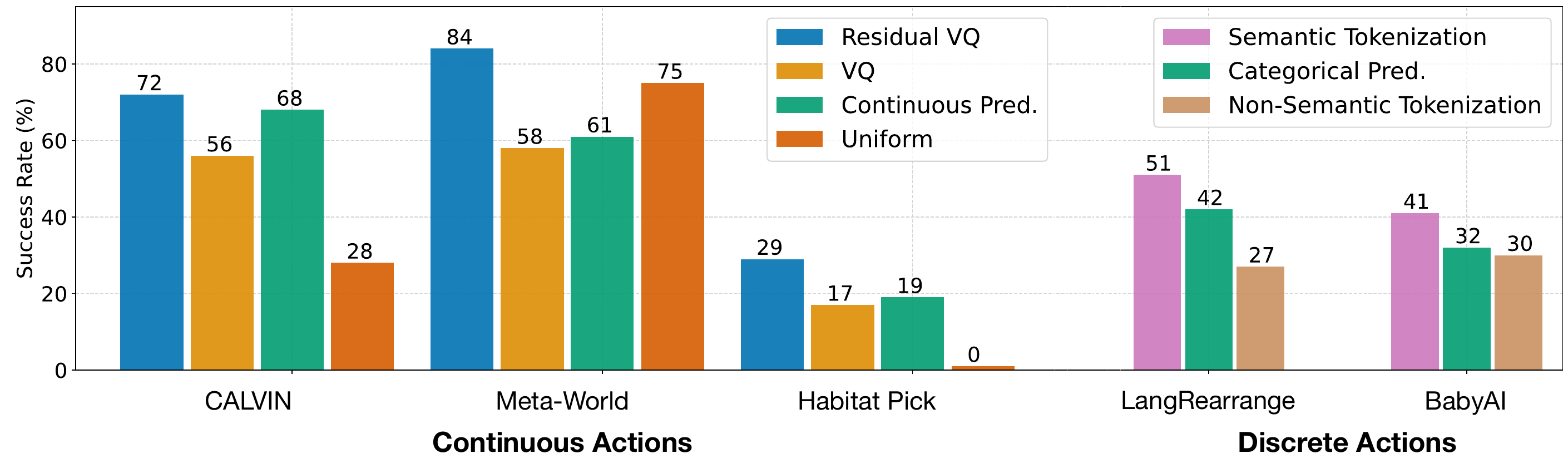}
  \caption{
    Comparing \asas for continuous and discrete action spaces across 5 environments. For continuous actions, the \rvq tokenization performs best. For discrete actions, \semlang performs best.
    Each bar gives the average over all tasks in the environment with the full breakdown in \Cref{sec:per-task}.
  }
  \label{fig:vlm-disc} 
  \label{fig:vlm-cont} 
\end{figure*}

\section{Experiments}
\label{sec:experiments} 

\subsection{Experimental Settings}
\label{sec:exp-settings} 
We study adapting MLLMs for action across a variety of environments with different embodiments and action spaces. All environments provide RGB visual observations and a natural language instruction specifying the goal to achieve. 
We provide the important environment details below and defer complete details to \Cref{sec:env-details}.

\textbf{CALVIN}~\cite{mees2022calvin}: This manipulation benchmark tests the ability of a tabletop robot to interact with an object to complete a natural language instruction. 
The continuous actions specify 6DoF end-effector control and the binary gripper state. 
The observation is a $ 200 \times 200$ RGB image from a fixed-position camera. 
We use the $ ABC \rightarrow D$ split of the benchmark with 34 tasks, and the agent is evaluated on unseen instruction phrasings and table background.

\textbf{\metaworld}~\cite{metaworld}: We use the ML-45 version of this tabletop manipulation benchmark which has 45 tasks.
The action space is continuous control specifying 3DoF end-effector translation and the continuous gripper state.
The observations are $ 200 \times 200 $ RGB images from a fixed camera. 
The agent is evaluated on unseen object and robot starting states.

\textbf{Habitat Pick (HabPick)}~\cite{szot2021habitat}: A mobile manipulation robot must pick up an object specified by name from a receptacle. 
The continuous actions specify the 7DoF relative joint positions of the arm, the 2D base velocity, and the gripper state. 
The observations are $ 336 \times 336$ RGB images from the robot's egocentric head camera.
The instruction specifies the name of the object type to pick up.
The evaluation distribution is on unseen houses and new arrangements of objects.

\textbf{BabyAI} \cite{babyai_iclr19}: BabyAI is a grid world task where an agent navigates and interacts with objects to complete an instruction.
The discrete action space consists of navigation and interaction actions. 
The observation is a $ 200 \times 200$ RGB top-down view.
We use the five tasks from \cite{carta2023grounding}, and we report generalization to instructions rephrased with synonyms. 

\textbf{Language Rearrangement (LangR)}~\cite{szot2023large}: A mobile manipulation robot must rearrange objects to complete instructions like “store all the fruit in the fridge”. The discrete actions are 70 high-level skills to interact with objects and navigate.
The observation is a $ 336 \times 336$ RGB head camera.
Evaluation instructions test generalization to unseen houses and 10 unseen instruction datasets measuring paraphrastic robustness and behavior generalization.

In all environments, we report the success rate as the fraction of episodes in which the agent completed the language instruction. 
We use the success criteria provided by each environment. 
We train a policy per action adapter for each environment and report the generalization performance in the main text. 
When reporting a single success rate per environment, it is the success averaged between all evaluation episodes containing all tasks.
We give the full per-task breakdown for results in \cref{sec:per-task}.
\calvin, \metaworld, \habpick, and BabyAI provide expert demonstrations succeeding at the task. \calvin has $ 17.9k$ from humans, \metaworld $ 22.5k$ from a scripted policy,  \habpick $ 6.7k$ generated from an RL policy, and BabyAI $ 5k$ from a scripted policy. 
Full details on the train and evaluation setups per environment are in \Cref{sec:env-details}.

We train with supervised finetuning for \calvin, \metaworld, \habpick, and BabyAI.
We train with reinforcement learning on \langR. 
As described in \cref{sec:training} we train $ \approx 140M$ parameters with LoRA~\cite{hu2021lora}.
We use the AdamW optimizer~\cite{loshchilov2017decoupled} with a learning rate of $3\mathrm{e}^{-4}$, a warmup period of $ 10 \%$ of the total number of training steps, and cosine learning rate decay to $ 0$ by the end of training. 
For RL, we use PPO~\cite{schulman2017proximal}.
For the learned tokenization action space adapters, we, by default, use a codebook size of 512 with 512 dimensions per codebook element.
Complete hyperparameter and policy details are in \Cref{sec:method-details}.

\begin{figure*}[t]
  \centering
  \begin{subfigure}{0.24\textwidth}
    \includegraphics[width=\textwidth]{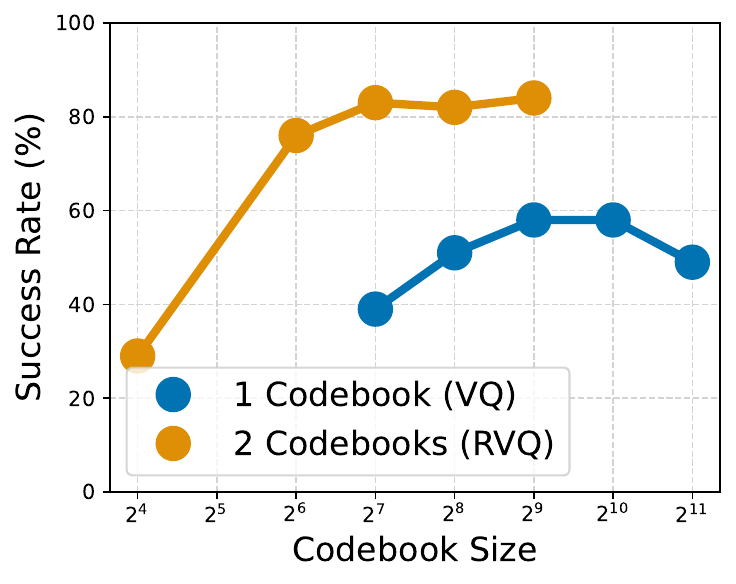}
    \vspace{-15pt}
    \caption{\# Codes: Success}
    \label{fig:vq-analysis:dim}
  \end{subfigure}
  \begin{subfigure}{0.24\textwidth}
    \includegraphics[width=\textwidth]{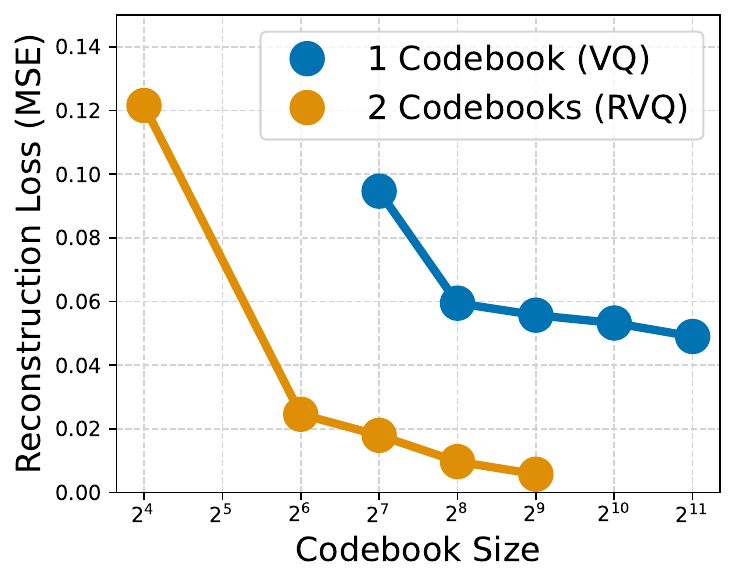}
    \vspace{-15pt}
    \caption{\# Codes: Recon.}
    \label{fig:rec:cbd} 
  \end{subfigure}
  \begin{subfigure}{0.24\textwidth}
    \includegraphics[width=\textwidth]{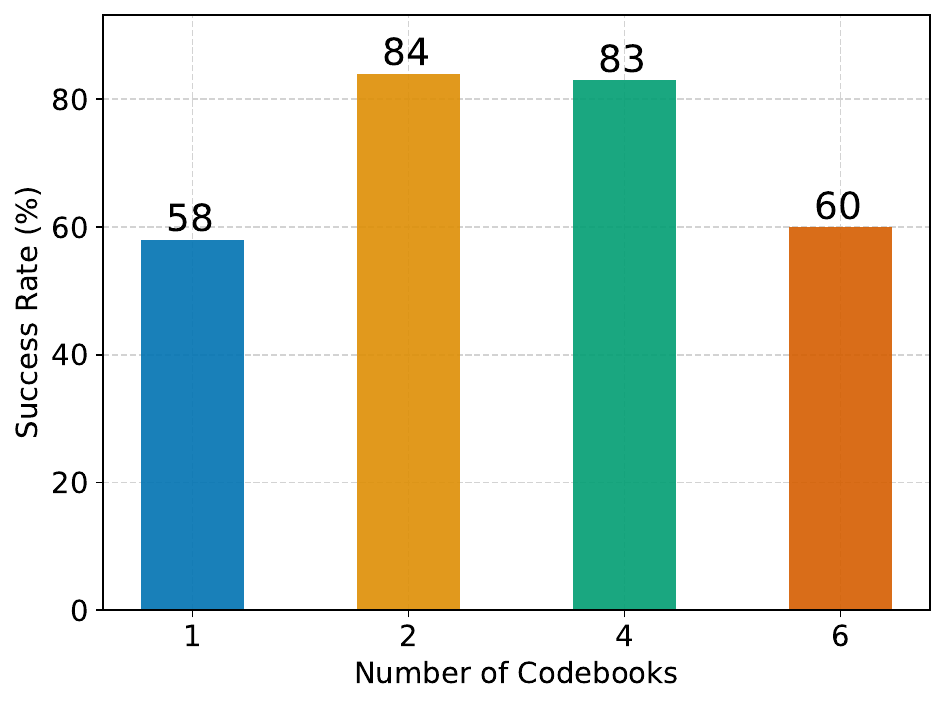}
    \vspace{-15pt}
    \caption{\# Codebooks: Success}
    \label{fig:vq-analysis:count}
  \end{subfigure}
  \begin{subfigure}{0.24\textwidth}
    \includegraphics[width=\textwidth]{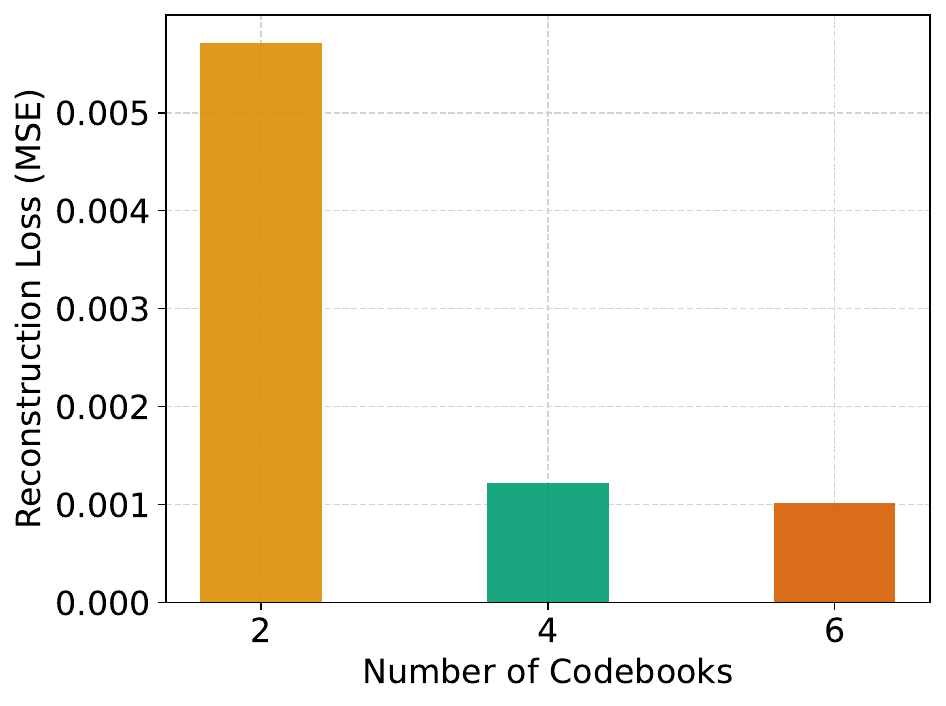}
    \vspace{-15pt}
    \caption{\# Codes: Recon.}
    \label{fig:rec:ncbs} 
  \end{subfigure}
  \vspace{-5pt}
  \caption{
    (a,b) show the effect of the number of codes in the codebook for \rvq and \vq on final policy success rate (see (a)) and reconstruction on unseen action trajectories in \metaworld (see (b)).
    (c,d) show the effect of number of codebooks on final policy success rate (see (c)) and action reconstruction (see (d)). All metrics are computed on \metaworld.
  }
  \label{fig:vq-analysis} 
\vspace{-0.5cm}
\end{figure*}

\subsection{Continuous Action Space Adapter Comparison}
\label{sec:action-adapter}

We first study adapting \vlms through \unif, \mlp, \vq, and \rvq  action space adapters for the continuous action environments \calvin, \metaworld and \habpick. 

\textbf{\rvq is the best performing continuous action \asa.}
The results in \Cref{fig:vlm-cont} show that the \rvq action adapter consistently outperforms all other \asa approaches across all environments. 
While \mlp is the second best performing method on all tasks, except on \metaworld, \rvq outperforms it by a $12\%$ average absolute difference. 
One hypothesized reason for this is that \mlp only learns unimodal distributions of actions, which hurts performance when learning from diverse demonstrations~\cite{lee2024behavior,shafiullah2022behavior,cui2022play}. 
Another potential reason is the tokenization from \rvq allows the \vlm to better leverage its existing knowledge, whereas the \mlp \asa requires training a new MLP network from scratch. 

\unif performs poorly on the majority of the tasks, where \rvq outperforms on average by a $ 27\%$ absolute increase. 
A reason for this is that the \unif discretization can fail to accurately represent the continuous actions. 
The performance of \unif is also closely related to the action dimension. In \metaworld with 4 action dimensions, \unif performs well. However, \unif suffers with the 7 action dimensions in \calvin and the 10 action dimensions in \habpick. 

\rvq also outperforms \vq by a $18\%$ absolute difference averaged over all environments. 
This is due to \vq having worse action reconstructions than \rvq.
In \metaworld, both \rvq and \vq policies reach a similar cross-entropy loss on holdout trajectories during finetuning.
However, on this same data, \rvq has a reconstruction mean squared error (MSE) of $ 0.005$ while \vq has a 10x higher reconstruction MSE of $ 0.05$. 
Increasing the \vq codebook size does not close this gap.
We vary the \vq codebook size in powers of $ 2$ from $ 2^{7}$ to $ 2^{11}$. 
\Cref{fig:rec:cbd} shows the \vq reconstruction loss decreases with larger codebooks but does not even close the gap to the $ 2^{7}$ \rvq codebook size. 
This poor reconstruction manifests in poor downstream policy performance as demonstrated by \Cref{fig:vq-analysis:dim} where policies trained with the \vq \asa plateau in success rate at codebook size $ 2^{9} $. 
\vq policies even decrease in performance at codebook size $ 2^{11}$, potentially due to overfitting to the large codebook.

\begin{figure*}[t] 
  \centering
    \includegraphics[width=0.8\textwidth]{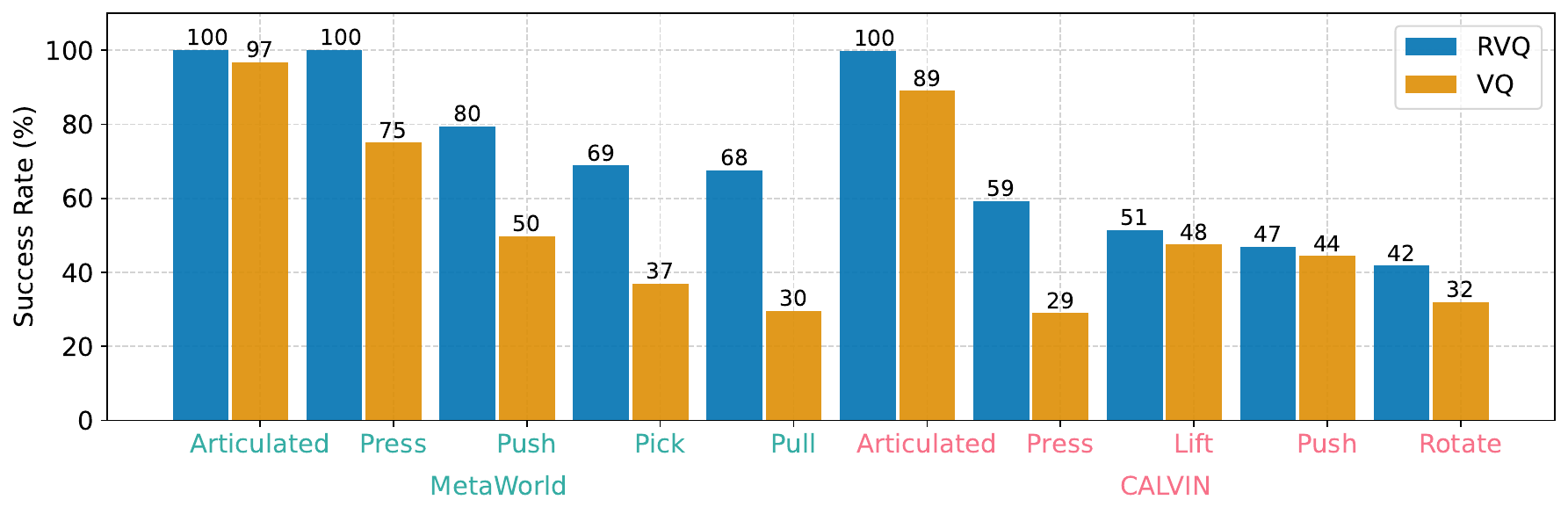}
    \caption{
    \rvq and \vq success per-task grouping (defined in Supp.~\ref{sec:task-groupings}) on \calvin and MetaWorld.
    }
    \label{fig:breakdown} 
\vspace{-10px}
\end{figure*}

We further characterize the performance of \rvq and \vq in \Cref{fig:breakdown} by breaking down the performance per task group in \metaworld and \calvin. The task groups, which are fully listed in \Cref{sec:task-groupings}, correspond to tasks with related required behaviors. 
Both \rvq and \vq do similarly on ``articulated" object interactions (like opening drawers or doors).
These tasks require less precise control since many contact points on the articulated link and broad pushing or pulling behavior can achieve the desired behavior. 
On the other hand, \rvq outperforms \vq on ``pressing"  tasks that require pushing a button. 
These tasks require more precise control since the agent needs to push the button all the way to a desired state.
\vq often reaches the button but fails to press it all the way. 
The same is also true of other precise control tasks like picking, pulling, and rotating. 

A potential explanation of RVQ's success can be attributed to adaptive localization of the model's errors, similar to prior work in residual reinforcement learning~\cite{johannink2019residual} and Bellman error bases~\cite{parr2008analysis}.

\textbf{A sufficient codebook size and number of codebooks are necessary for \rvq.}
In \Cref{fig:vq-analysis:dim}, we show that \rvq policy performance improves in performance with a larger codebook size in \metaworld. 
Notably, \rvq performs poorly at $ 29\%$ success rate with codebook size $ 16$ compared to $ 84\%$ success at codebook size $ 512$. 
These observations also align with the codebook size decreasing reconstruction error in \Cref{fig:rec:cbd}.
In \Cref{fig:vq-analysis:count}, we compare the effect of the number of codebooks on performance. 
As earlier discussed with the performance of \vq, one codebook results in poor action reconstruction and, thus, bad policy performance. 
However, increasing the number of codebooks too much to 6 also hurts performance despite decreasing reconstruction loss.
Likewise to the finding that \unif performs poorly with larger action dimension since there are more tokens per action, increasing the number of codebooks also hurts policy learning. 

\textbf{\rvq tokens transfer to new tasks}. 
We take the model trained on the 45 \metaworld tasks and finetune it on 5 unseen tasks. We collect 50 demonstrations for per task and finetune the policy on all task data. 
We use the same \rvq \asa trained only on data from the 45 tasks. 
\Cref{fig:mw-adapt} shows the success rate of adapting \rvq compared to an \mlp \asa. 
\rvq outperforms \mlp across all tasks, achieving a $ 50\%$ vs. $ 20\%$ overall success rate. 
This demonstrates the \rvq tokens are flexible enough to be applied to new tasks.

\textbf{The gains from \rvq are unique to \vlms.}
Next, we analyze the unique interaction between the \rvq tokens and the \vlm policy. 
While we demonstrated that the \rvq \asa performs best, is this improvement due to the \vlm being able to leverage these new tokens or the added action representation ability from the separately trained \rvq decoder? 
To test this, we compare to two policy architectures that do not use LLMs: 
\begin{itemize}
[itemsep=1pt,topsep=0pt,parsep=0pt,partopsep=0pt,parsep=0pt,leftmargin=*]
  \item \textbf{\scratch}: This is the same architecture as the \vlm-based policy, but with a smaller 300M parameter non-pretrained transformer.

  \item \textbf{\rt}: This method uses a ResNet visual encoder, pretrained Flan~\cite{wei2021finetuned} language embedding and decoder transformer-based policy. The entire policy is trained from scratch. This method is inspired by RT-1~\cite{brohan2022rt}, which does not have publicly released code.
\end{itemize}
\Cref{table:llm-cmp} compares the effect of \mlp versus \rvq \asas on \calvin,  \metaworld and \habpick for these three policy policy architectures.
As already established for the \vlm, \rvq is consistently better than \vq.
However, for the same policy architecture trained from scratch, \rvq can hurt the performance over \mlp. 
In \calvin the success drops $ -7\%$ and in \metaworld the performance drops $ -15\%$.
This highlights that \vlm can leverage its existing knowledge about sequencing language tokens to sequencing action tokens.
However, we find that for the smaller \rt policy network, the \rvq \asa consistently helps, which we hypothesize is because the added \rvq network and separate training help compensate for the lack of policy network capacity.
We also note that \rvq may more consistently outperform \mlp on demonstrations that explicitly contain multimodal action sequences~\cite{lee2024behavior,shafiullah2022behavior,cui2022play}.

\subsection{Discrete Action Adapter Comparison}
\label{sec:disc-asa} 
\begin{figure*}[b]
  \centering
  \begin{subfigure}{0.4\textwidth}
    \includegraphics[width=\textwidth]{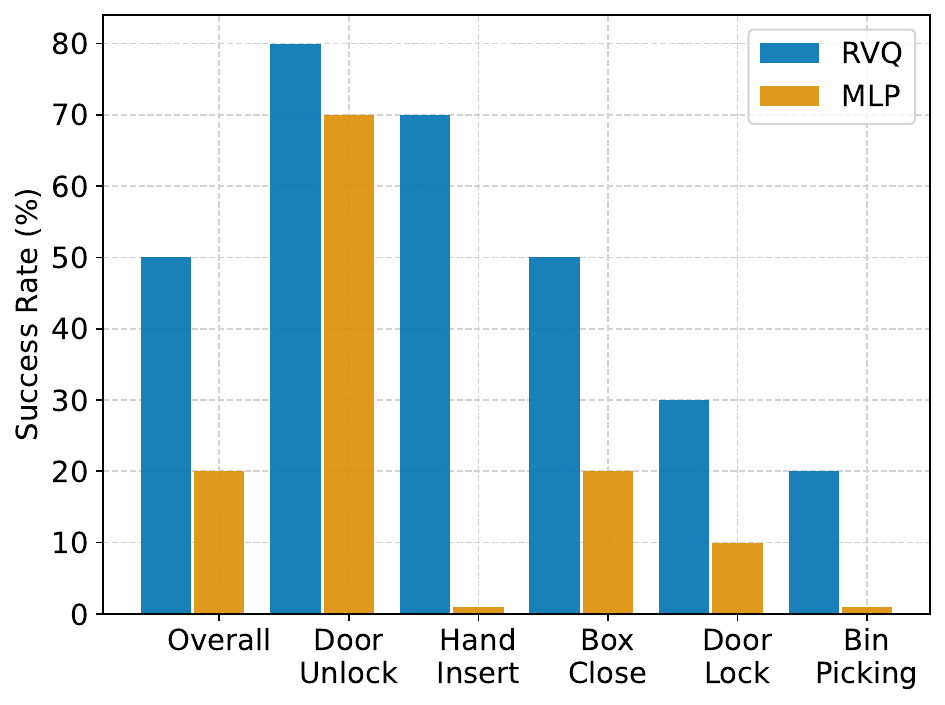}
    \caption{Finetuning on Holdout Tasks}
    \label{fig:mw-adapt}
  \end{subfigure}
  \begin{subfigure}{0.38\textwidth}
    \includegraphics[width=\textwidth]{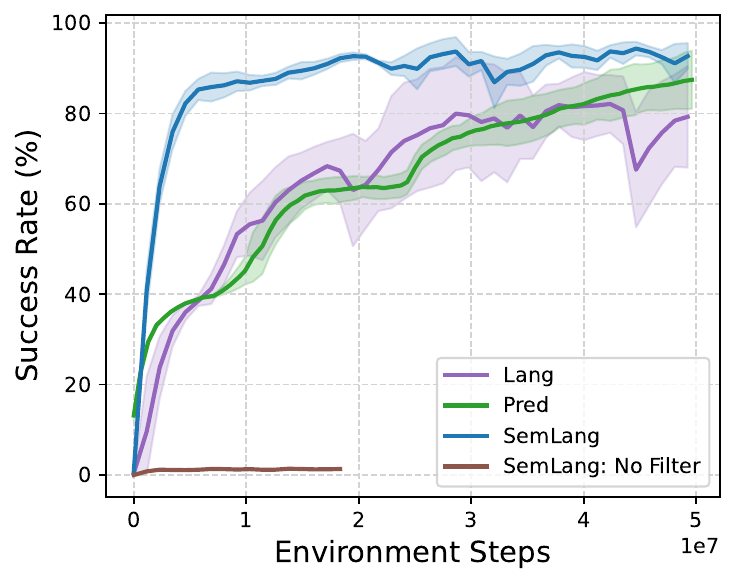}
    \caption{
      \langR RL.
    }
    \label{fig:langR-rl}
  \end{subfigure}
  \caption{
    (a) Adapting to 5 holdout tasks from \metaworld ML-45 with 50 demos per task using the fixed \rvq tokenization.
    (b) RL training curves in \langR comparing the \asas and utility of the token filter. 
    Displayed are averages over 2 seeds with the shaded area as the standard deviation between seeds.
    \semlang learns faster than other \asas and the token filter is crucial.
  }
\end{figure*}

\textbf{\semlang performs the best.}
In \Cref{fig:vlm-disc}, \semlang outperforms the next best \asa (\mlp), by $ 9\%$ on \langR and $ 8\%$ on BabyAI.
\semlang performs especially well on tasks with explicit high-level language actions in \langR (e.g., ``pick apple'') where prior work has shown text-only LLM policies achieve non-zero success~\cite{szot2023large}. 
\semlang also does well on the BabyAI tasks with discrete low-level actions like ``move left". 
Additionally, \lang performs the worst in both environments, achieving $ 14\%$ lower success on \langR and $ 11 \%$ lower on BabyAI than \semlang.
We hypothesize this is because the \vlm has to repurpose its knowledge to leverage these newly assigned action tokens, whereas a newly initialized \mlp allows extracting this knowledge from the \vlm hidden state.

\textbf{\semlang enables sample efficient RL.} 
In \Cref{fig:langR-rl}, we compare the RL training curves for the \asas in \langR.
In addition to helping with better generalization, \semlang also enables sample efficient RL training.
\semlang converges in training performance after just $ 20M$ training samples, whereas  \mlp requires up to $ 70M$ steps to fully converge. 

\textbf{Token filter is crucial for language-based action spaces.} 
In \Cref{fig:langR-rl}, we show the training of \semlang without the token filter, which restricts policy outputs to only valid action token sequences. 
Without the token filter, \semlang is unable to learn in the large text action space.

\begin{table*}[t!] 
  \centering
  \input{tables/llm_cmp.tex}
  \caption{
    Comparing the effect of the \rvq action space adapter on the success rate of non-LLM based policies. Red indicates \textcolor{lightred}{\rvq hurts over \mlp} and green indicates \textcolor{lightgreen}{\rvq helps over \mlp}. \rvq typically has a negative impact on the \scratch policy, and helps the smaller \rt policy.
  }
  \label{table:llm-cmp}
\vspace{-0.2cm}
\end{table*}
\subsection{Empirical Comparison to Prior Work}
The contributions of this work are an empirical analysis of \asas under controlled settings on various embodied environments.
Direct comparisons to prior work are challenging due to different training algorithms, policy architectures, or assumptions about input modalities.
Regardless, in this section, we seek to contextualize our \rvq and \semlang \vlm results against prior work.
In \metaworld, to the best of our knowledge, \rvq at $ 84\%$ success on ML-45 sets a new state-of-the-art result, compared to $ 79\%$ from DualMind~\cite{wei2023imitation}. 
In \calvin, \rvq at $ 72\%$ success underperforms a similar work RoboFlamingo which achieves $ 82\%$ success on the $ ABC \rightarrow D$ split. 
However, RoboFlamingo uses a different \vlm and uses an additional gripper camera input.
In \langR, \semlang sets a state-of-the-art result with $ 51\%$ success compared to $ 42 \%$ from LLaRP~\cite{szot2023large}.
In BabyAI, \semlang at $ 40\%$ success rate underperforms GFlan~\cite{carta2023grounding}, which achieves $ 55\%$ success. However, we use RGB visual observations, while GFlan operates from a compact, ground truth language state description.
In \Cref{sec:prior-work-exp}, we compare these differences in more detail.

%% file: tables/llm_cmp.tex
\begin{tabular}{rllll}
\toprule
 & \textbf{\vlm: \mlp $ \!\rightarrow\!$ \rvq} & \textbf{\scratch: \mlp $ \!\rightarrow\!$ \rvq} & \textbf{\rt: \mlp $ \!\rightarrow\!  $ \rvq} \\
\midrule
  \textbf{Calvin} &  $ 68 \rightarrow 72 $ \color{lightgreen}{(+4)} &  $ 50 \rightarrow 43,$ \color{lightred}{(-7)} &  $ 35 \rightarrow 36,$ \color{lightgreen}{(+1)} \\
\textbf{Metaworld} &  $61 \rightarrow 84$ \color{lightgreen}{(+23)} &  $ 71 \rightarrow 56,$ \color{lightred}{(-15)}&  $ 27 \rightarrow 38,$    \color{lightgreen}{(+11)}\\
\textbf{Habitat Pick} &  $19 \rightarrow 29$ \color{lightgreen}{(+10)} &  $ 21 \rightarrow 25,$  \color{lightgreen}{(+4)}&  $ 18 \rightarrow 20,$ \color{lightgreen}{(+2)} 
\end{tabular}

%% file: sections/conclusion.tex
\section{Limitations and Conclusion}
\label{sec:conclusion} 
\vspace{-0.2cm}
In this work, we studied various action space adapters (\asas) across a variety of embodiments, action spaces, and environments. We provide a generalization of prior works through the lens of action space adaptors, and for both discrete and continuous action spaces demonstrate designs that we show can leverage the knowledge within the \vlm.
Our findings conclude that for continuous actions, it is best to learn action tokens that accurately model the action distribution, while for discrete actions, it is best to reason over semantic language descriptions of actions.
We verify these ideas across \taskcount embodied AI tasks in 5 diverse environments.

A limitation of our work is all our analysis is under a single \vlm (\llava). Another limitation is that \rvq, the best performing \asa in continuous action spaces, requires collecting demonstrations to train the VQ model. Our analyses are also under only a single LoRA training setting. Future analyses can explore different base \vlms under different training regimes like full LLM finetuning. While our investigation of ASAs enables connecting a MLLM to various action spaces, the performance of these methods is still subpar for real-robot deployment where high success and safety are critical. MLLMs with the best ASA still struggle on simple environments like BabyAI, only achieving 40\% success rate. Further work is needed to improve the performance of these methods for real-world usage. Our investigation also only studies adapting MLLMs through behavioral cloning or on-policy RL. Future work can investigate if the choice of ASA varies when adapting the MLLM with other learning algorithms such as off-policy RL or offline RL.

%% file: supp/prior_work.tex
\section{Prior Work Comparison}
\label{sec:prior-work} 

In this section we expand on the differences between the prior work in action space adaptation mentioned in \Cref{sec:related-work} and our investigation. 
\Cref{table:prior-work} compares our investigation to prior work along several key dimensions. We emphasize that unlike prior works, ours studies a variety of action space adapters under a greater diversity of environments. 

\begin{table}
\resizebox{\columnwidth}{!}{
  \begin{tabular}{lp{6cm}p{3cm}p{5cm}llc}
    \toprule
    Work & Environments & Best \asa & Other \asas Studied & Action Space Types & Training & Using LLM/VLM? \\
    \midrule
    RoboFlamingo~\cite{li2023vision} & CALVIN & MLP & - & Continuous & BC & \greencheck \\
    Lamo~\cite{shi2023unleashing} & Franka Kitchen, Atari, MuJoCo & MLP & - & Continuous, Discrete & Offline-RL & \greencheck \\
    GFlan~\cite{carta2023grounding} & BabyAI & Sem Lang (scoring) & - & Discrete & Online RL & \greencheck \\
    RT-2~\cite{brohan2023rt} & Internal & Uniform & - & Continuous & BC & \greencheck \\
    LLaRP~\cite{szot2023large} & Language Rearrangement & MLP & - & Discrete & Online-RL & \greencheck \\
    VQ-BeT~\cite{lee2024behavior} & PushT, Multimodal Ant, BlockPush, Franka Kitchen, nuScenes, PlayKitchen & RVQ+MLP & - & Continuous & BC & \redx \\
    Ours & Language Rearrangement, Baby AI, MetaWorld, CALVIN, Habitat Skills & RVQ/ Sem-Lang & MLP, VQ, Uniform, Non-Sem, Non-Sem Comp &Continuous, Discrete & Online-RL, BC & \greencheck \\
    \bottomrule \\
  \end{tabular} 
}
\caption{
  Comparing our investigation to prior work. 
  Prior work typically analyzes a single action adapter in a single environment. 
  We study a variety of action adapters across a variety of environments. 
}
\label{table:prior-work} 
\end{table}

\subsection{Empirical Comparison to Prior Work}
\label{sec:prior-work-exp} 
We report performance on standard benchmarks which prior work has also extensively studied. However, even within the benchmarks there are differences in training algorithms and sensor input assumptions that make direct comparison to prior work difficult. Regardless of these differences, we study different \asas for \vlms in a consistent experimental setting. We also describe differences between the empirical setups of ours and prior works that perform well on these benchmarks.

\textbf{\metaworld} (\vlm+\rvq 84\% success rate on ML-45): To the best of our knowledge, our $ 84\%$ is the highest reported on \metaworld ML-45 so far. \citet{anand2021procedural} operates under similar sensor assumptions and achieves $ 77\%$ success with MuZero~\cite{schrittwieser2020mastering}. DualMind~\cite{wei2023imitation} achieves $ 79\%$ success rate on ML-45 and outperforms other generalist agents like Gato~\cite{reed2022generalist}. However, DualMind uses privileged simulator information about the joint states and object positions while we only use RGB visual observations.

\textbf{\calvin} (\vlm+\rvq 72\% success rate): RoboFlamingo achieves a higher $ 82\%$ success rate on the same $ ABC \rightarrow D$ task. However, RoboFlamingo uses the OpenFlamingo VLM while we use \llava. RoboFlamingo use the gripper and fixed camera while we only use the fixed camera. 
More recent work like 3D Diffuser Actor~\cite{ke20243d} practically solves the $ ABC \rightarrow D$ task, achieving $ 96 \%$ success rate. However, this work uses depth inputs, and a diffusion model policy that predicts keypoints for the end-effector rather than underlying actions. 
Our work uses only RGB visuals, uses a \vlm policy and predicts relative end-effector poses rather than keypoints.

\textbf{\langR} (\semlang 51\% success rate): This outperforms the prior highest reported number of $ 42 \%$ on the overall evaluation set from LLaRP~\cite{szot2023large}. 

\textbf{BabyAI} (\semlang 40\% success rate): GFlan~\cite{carta2023grounding} achieves $ 55\%$ success on the same evaluation split. However, the GFlan policy takes as input a ground truth language description of the state, while our policies take as input a $ 200\times 200$ RGB top down rendering of the environment. GFlan also trains the policy with reinforcement learning while we train with supervised learning. 

%% file: supp/exps.tex
\begin{figure*}[!h]
  \centering
  \includegraphics[width=\textwidth]{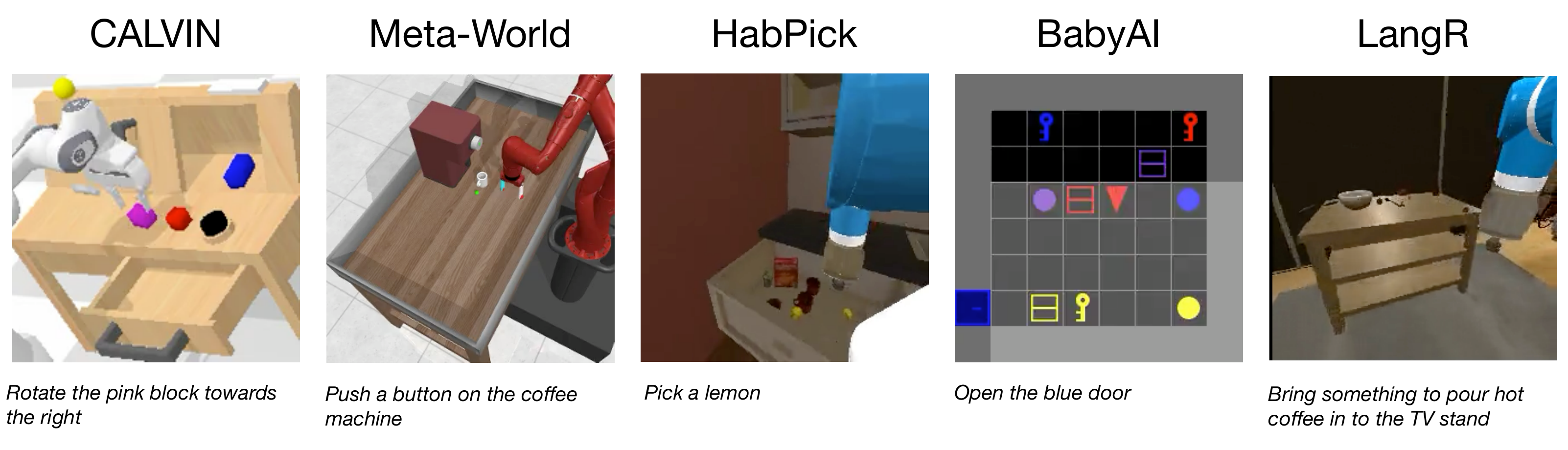}
  \caption{
    Visualizations of the environments we study. The top row shows an observation in the environment. The bottom row shows the associated instruction in that episode.
  }
  \label{fig:envs}
\end{figure*}

\section{Environment Details}
\label{sec:env-details}
An overview of the environments is visualized in \Cref{fig:envs}. This figure visualizes the training observations input to the agent. 
We run experiments on 5 environments, and each environment in turn consists of multiple tasks.
We arrive at the task count of \taskcount in the main paper through 45 tasks in \metaworld, 34 in \calvin, 20 in \habpick where we count each object goal as a different task, 10 in \langR for each of the evaluation splits, and 5 in BabyAI.
The task count for \langR is conservative since technically it consists of 282 instruction templates, each of which corresponds to a distinct task and goal.

\subsection{\metaworld}
\textbf{Tasks}: We use the ML-45 benchmark from \metaworld~\cite{metaworld}. Each of the 45 tasks are specified with a fixed language instruction. We use the task descriptions from Appendix Section A of \citet{metaworld}.

\textbf{Observation Space}: $ 200 \times 200$ RGB images from a fixed camera position. 
To render the visual observations, we only use the ``corner4" camera position as this gives an unobstructed view of the robot in most of the tasks. 

\textbf{Action Space}: 4DoF continuous control of the arm and gripper. The first 3 dimensions specify the relative end-effector translation. The last dimension specifies the desired gripper state.

\textbf{Training}: We use 40 start and goal configurations for each of the tasks. 
We generate 500 demonstrations for each of the 45 tasks. We use the scripted policies from~\citet{metaworld}. At each step we add Gaussian noise $ \mathcal{N}(0, 0.1)$ to the actions produced by the scripted policy before executing it in the environment.  We generate 500 successful trajectories per task, resulting in $ 45 \cdot 500 = 22.5k$ total trajectories.

\textbf{Evaluation}: We evaluate performance on 10 unseen start and goal configurations for each of the 45 tasks.
So in total, we evaluate on 450 unseen configurations and report the average performance over these 450 episodes.

\subsection{\calvin}
\textbf{Tasks}: We use the \calvin $ ABC \rightarrow D$ dataset split. 

\textbf{Observation Space}: $ 200 \times 200$ RGB observations from the fixed camera view. 

\textbf{Action Space}: 7DoF continuous control of the arm and gripper. The first 6 dimensions specify the relative position and rotation of the end-effector. The final dimension is a binary indicator for if the gripper should be open or closed.
We hold out $ 1024$ subsequences of the policy context length from these trajectories for reporting validation performance during the SFT process. 

\textbf{Training}: 
We use the 17,871 demonstrations provided in the CALVIN $ ABC \rightarrow D$ dataset. 
These demonstrations are in 3 different table backgrounds. 
This also includes 1,088 demonstrations for validation. 

\textbf{Evaluation}: 
We report performance on the $ D$ split. 
This evaluation scene is a different color than that encountered during training. 
All the start positions and goals are also different.
Many of the language instructions are also unseen from training.
We report the average performance over the 1,000 evaluation sequences. 
We report the success of the first task completed in the sequence.

\subsection{\Habpick}
\textbf{Tasks}: We use the same Pick task as in Habitat 2.0 Geometric Goal object rearrangement~\cite{szot2021habitat,habitatrearrangechallenge2022}, except we provide the agent the name of the object to rearrange rather than the starting coordinates of the object and increase the observation resolution. 
The task is successful when the agent picks up the object and returns the end-effector within a fixed offset to a ``resting position" in front of the robot.
The task ends in failure if the agent excessively collides with the scene, drops the object, or picks up the wrong object. 
The agent starts within 2 meters of the object and facing towards the receptacle but with random noise $ \mathcal{N}(0, 1.57)$ applied to the direction of facing directly at the receptacle.
The maximum number of steps per episode is 300 steps.

\textbf{Observation Space}: A $ 336 \times 336$ head-mounted RGB camera. 

\textbf{Action Space}: The action space is 10DoF control of the arm, base and gripper. The first 2 dimensions control the linear and angular velocity of the base. The next 7 dimensions control the relative joint offsets of the arm. The final dimension controls whether the suction gripper is engaged or not. 

\textbf{Training}: 
We first train a privileged policy with RL to complete the task. This policy takes as input the egocentric depth image and the ground truth position of the target object to pick up. We collect $ 20k$ successful trajectories.

\textbf{Evaluation}: 
We evaluate on the test episodes from \citet{habitatrearrangechallenge2022} which are $ 1,000$ episodes in unseen home layouts.

\subsection{BabyAI}
\textbf{Tasks}: 
The tasks all occur in a $ 6 \times 6$ grid populated with interactable objects. 
We use the task definitions from \citet{carta2023grounding}. This consists of the following 5 instruction templates: ``Go to $ <$object$>$", ``Pick up $<$object$>$", ``Put $<$object A$ >$ next to $ <$object B$>$,", ``Pick up $<$object A$>$ then go to $<$object B$>$ and Go to $<$object B$>$ after pick up $<$object A$>$", ``Unlock $<$door$>$". 
The maximum number of steps per episode is 50 steps.

\textbf{Observation Space}: $ 200 \times 200$ RGB observation as a top down of the $ 6 \times 6$ scene.
Note this is a more challenging observation space than prior gridworld navigation tasks that provide the current view as a compact entity specific array~\cite{babyai_iclr19} or by a language description~\cite{carta2023grounding}. 

\textbf{Action Space}: 
The action space consists of 6 actions consisting of: turn left, turn right, move forward, pick, drop and toggle. 

\textbf{Training}: 
We collect 1,000 demonstrations for each of the 5 templates. We randomly sample an instruction and starting state configuration for every demonstration. We use the expert planner from \citet{babyai_iclr19} to generate the demonstrations. 

\textbf{Evaluation}: 
We report performance on the unseen synonyms generalization test, described in Section 4.2 of \citet{carta2023grounding}.
We evaluate on 200 episodes per template type, giving 1000 total evaluation episodes.

\subsection{\LangR}
\textbf{Tasks}: An agent starts in an unseen house and must complete a rearrangement task from a language instruction. 

\textbf{Observation Space}: The agent has a $ 336 \times 336$ head-mounted RGB camera. We increase the camera resolution from $ 256 \times 256 $ in the original \langR task to match the input resolution of the \llava CLIP encoder.

\textbf{Action Space}: We use the same action space as from the original \langR benchmark \citet{szot2023large}. The agent can select between 70 high-level skills that include picking up objects by name, navigating to receptacles, placing on receptacles by name, and opening and closing receptacles by name.

\textbf{Training}: Since \langR does not provide any demonstrations and due to the emphasis on exploration in the problem, they are not readily obtainable, even with oracle planners. Therefore, we opt to train policies with reinforcement learning from the environment reward provided by the \langR task.

\textbf{Evaluation}: We evaluate on all 10 evaluation datasets from \langR consisting of 1,000 evaluation episodes on unseen scenes.

\subsection{Task Groupings}
\label{sec:task-groupings} 
In \Cref{sec:experiments} we breakdown the performance on CALVIN and MetaWorld for task groupings. Each of the task groupings consists of multiple tasks from the benchmark. We grouped tasks in the following way:

\textbf{MetaWorld}: 
\begin{itemize}[itemsep=0pt,topsep=0pt,parsep=0pt,partopsep=0pt,parsep=0pt,leftmargin=*]
\item Articulated: ``door-close", ``door-open", ``drawer-close", ``drawer-open", ``faucet-open", ``faucet-close", ``handle-press-side", ``handle-press", ``window-open", ``window-close"
\item Press: ``button-press-topdown", ``button-press-topdown-wall", ``button-press", ``button-press-wall", ``coffee-button"
\item Push: ``plate-slide", ``plate-slide-side", ``plate-slide-back", ``plate-slide-back-side", ``push-back", ``push", ``push-wall", ``stick-push", ``sweep-into", ``sweep", ``soccer", ``coffee-push"
\item Pick: ``assembly", ``basketball", ``dial-turn", ``disassemble", ``hammer", ``peg-insert-side", ``peg-unplug-side", ``pick-out-of-hole", ``pick-place", ``pick-place-wall", ``reach", ``reach-wall", ``shelf-place"
\item Pull: ``coffee-pull", ``handle-pull-side", ``handle-pull", ``lever-pull", ``stick-pull"
\end{itemize}

\textbf{CALVIN}: 
\begin{itemize}[itemsep=0pt,topsep=0pt,parsep=0pt,partopsep=0pt,parsep=0pt,leftmargin=*]
\item Articulated: ``move slider left", ``open drawer", ``close drawer", ``move slider right"
\item Press: ``turn off led", ``turn on led", ``turn on lightbulb", ``turn off lightbulb"
\item Lift: ``lift blue block slider", ``lift pink block table", ``lift red block slider", ``lift red block table", ``lift pink block slider", ``lift blue block table"
\item Push: ``push pink block right", ``push blue block right", ``push red block left", ``push pink block left", ``push red block right", ``push blue block left", ``push into drawer"
\item Rotate: ``rotate red block right", ``rotate red block left", ``rotate pink block left", ``rotate pink block right", ``rotate blue block right", ``rotate blue block left"
\end{itemize}

\section{Further Policy Details}
\label{sec:method-details} 

\subsection{Prompt Details}

In addition to inputting the task instruction to the LLM, we also format the instruction with a prompt. We base our prompt off the prompt used in \llava. For all continuous control tasks, we use the prompt template ``Prompt: control the robot. USER: $<$INSTRUCTION$>$ ASSISTANT: ". 
For discrete action space tasks, we describe the available actions to the agent in the prompt as well. For BabyAI, this is the prompt template ``Prompt: Control the red triangle to complete the instruction using left, right, forward, pick, drop and toggle. USER: $<$INSTRUCTION$>$ ASSISTANT: ". For \langR, this is the prompt template ``Prompt: You are a home robot assistant. Your possible actions are: pick object, place receptacle, nav receptacle, open receptacle, close receptacle, STOP. - Objects: ball, clamp, hammer, screwdriver, padlock, scissors, block, drill, spatula, knife, spoon, plate, sponge, cleanser, plum, pear, peach, apple, lemon, can, box, banana, strawberry, lego, cube, book, bowl, cup, mug, orange, lid, toy, wrench. - Receptacles: chair, black table, brown table, TV stand, sink, right counter, left counter, sofa, fridge, left drawer, right drawer, middle drawer. USER: $<$INSTRUCTION$>$ ASSISTANT: ".

\subsection{Action Space Adapter Details}
\label{sec:asa-details} 

We use the same \asa details between all environments. We detail the architecture and training decisions for the different \asas when applicable. 

\textbf{\vq}: Use a codebook size of 512 with 512 dimensions per codebook element. 
These 512 tokens are mapped to token indices $ 31000-31512$ from the \llama language modeling head.
The encoder and decoder networks for predicting the latent and decoding from the latent are 4 layer MLP networks with hidden size 2048 using ReLU activations. The \vq network is trained on the actions in the same dataset used to train the policy. 
The network is trained with MSE loss to reconstruct the original actions. 
We \vq network for 3 epochs over the dataset.

\textbf{\rvq}: Use all the same details as \vq, but with a Residual-VQ that uses 2 codebooks. 

\textbf{\mlp}: We use a 2 layer MLP network with a hidden size of 2048 and ReLU activations. We use this same MLP network architecture for discrete and continuous action space tasks. In the robot manipulation tasks, we also found it useful to include the robot proprioception as input to the MLP network and included this as input to the network layer. The robot proprioception consists of the robot robot joint angles and the gripper state. This \asa requires no separate training. 

\textbf{\unif}: In the tasks we consider, the actions are already normalized to be in $ [-1,1]$. We then create $ 512$ evenly spaced bins within this interval and assign each action dimension based on which bin it is within. Like with \vq, we assign the 512 tokens to indices $ 31000-31512$ from the \llama language modeling head. This \asa requires no separate training.

\textbf{\lang}: Starting from the same semantic tokenization as with \semlang, we remap each token to the token corresponding to a digit ``0" to ``9". Therefore, the token count per action is the same between \lang and \semlang, but the \lang action tokens have no semantic meaning being just digits.

\subsection{Training and Architecture Details}
\label{sec:training-details} 
\label{sec:architecture-details} 
\label{sec:hyperparams} 

We use all pretrained components from \llava. For the visual token downsampler, we use a 2 layer Perceiver network~\cite{jaegle2021perceiver} with 4 output latents and hidden size 4096. 

We detail the hyperparameters used for imitation learning in in \Cref{table:hyperparams}. We trained with the HuggingFace Transformers library~\cite{wolf2019huggingface}, PyTorch~\cite{paszke2019pytorch}, DeepSpeed~\cite{rasley2020deepspeed}. 
For reinforcement learning, we use learning rate $ 3e^{-4}$, $ 32$ steps per rollout. $ 18$ parallel environment workers per GPU, an entropy coefficient of $ 0.01 $, 2 epochs over the data batch per rollout, 6 PPO minibatches, a maximum gradient norm of 0.2 and $ \gamma = 0.99$. 

We train the \calvin, \metaworld and \habpick imitation learning results on a 4xA40 GPU setup. We train the \langR and BabyAI experiments on a 8xA100-80GB GPU setup. 

We train the LLM weights with LoRA and fine tune the entire \asa and downsampler module. For LoRA we use rank value 128, alpha parameter $ 32$ and dropout $ 0.1$.

\begin{table}
\rowcolors{1}{white}{lightgray}
  \centering
  \input{tables/hyperparams}
  \caption{
    Hyperparameters for all imitation learning experiments. Most hyperparameters are the same between environments but the number of training epochs, context length and batch size per GPU are adjusted to fit the need for history, environment dataset size and task complexity.
  }
  \label{table:hyperparams} 
\end{table}

\section{Qualitative Results}
\begin{figure*}
    \centering
    \begin{subfigure}[t]{0.49\textwidth}
        \includegraphics[width=\textwidth]{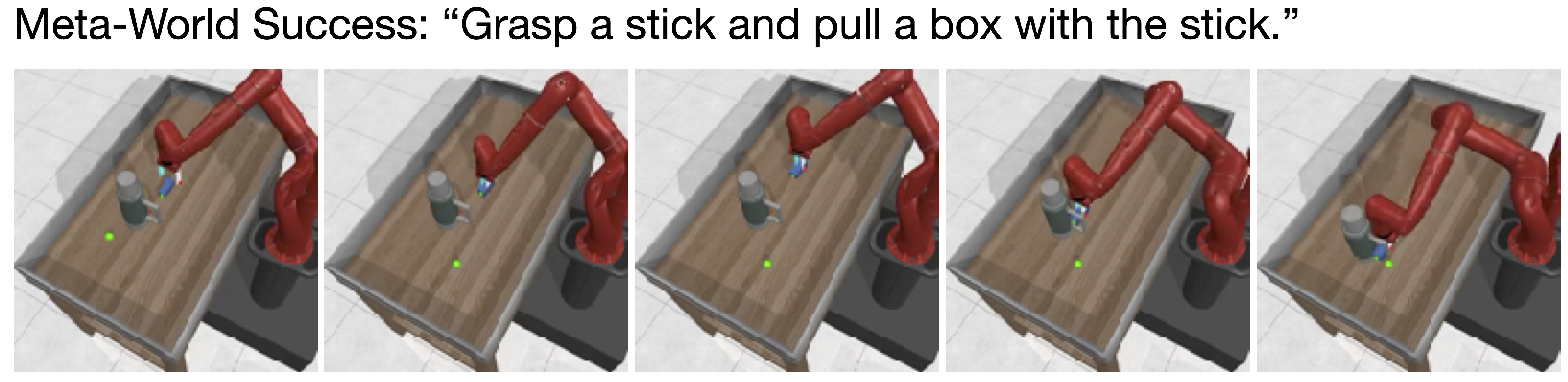}
        \caption{The robot successfully picks the stick and pushes the box to the goal position.}
    \end{subfigure}
    \begin{subfigure}[t]{0.49\textwidth}
        \includegraphics[width=\textwidth]{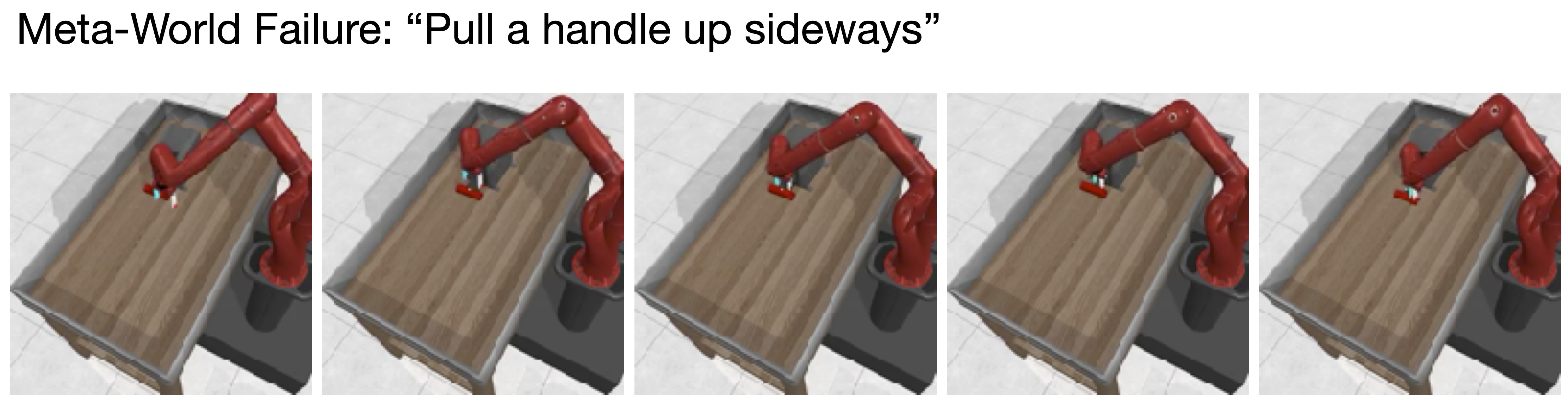}
        \caption{The robot only partially lifts the handle and fails to lift it up all the way.}
    \end{subfigure}
    \begin{subfigure}[t]{0.49\textwidth}
        \includegraphics[width=\textwidth]{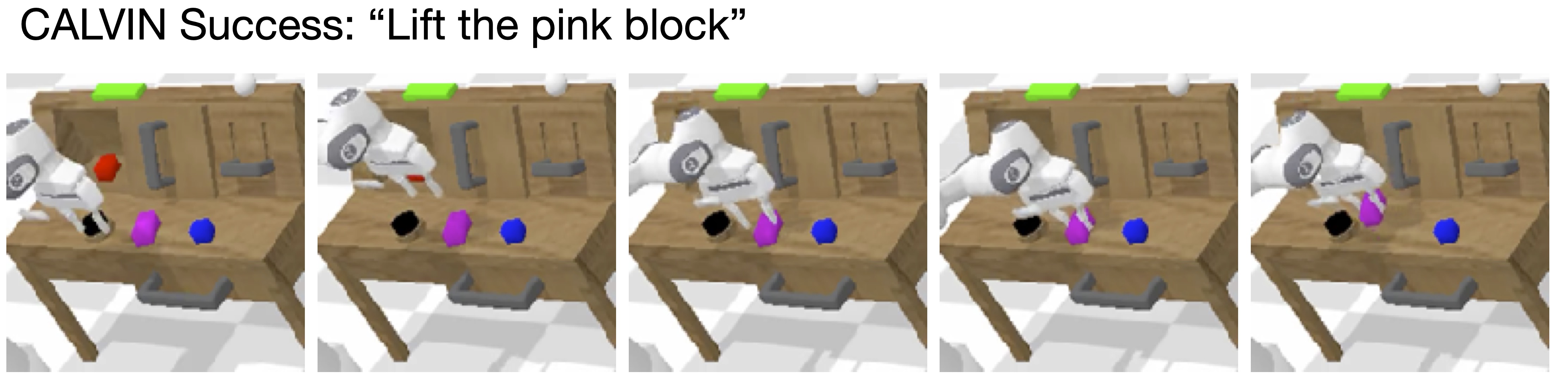}
        \caption{The robot grasps the pink block and lifts it to the goal height.}
    \end{subfigure}
    \begin{subfigure}[t]{0.49\textwidth}
        \includegraphics[width=\textwidth]{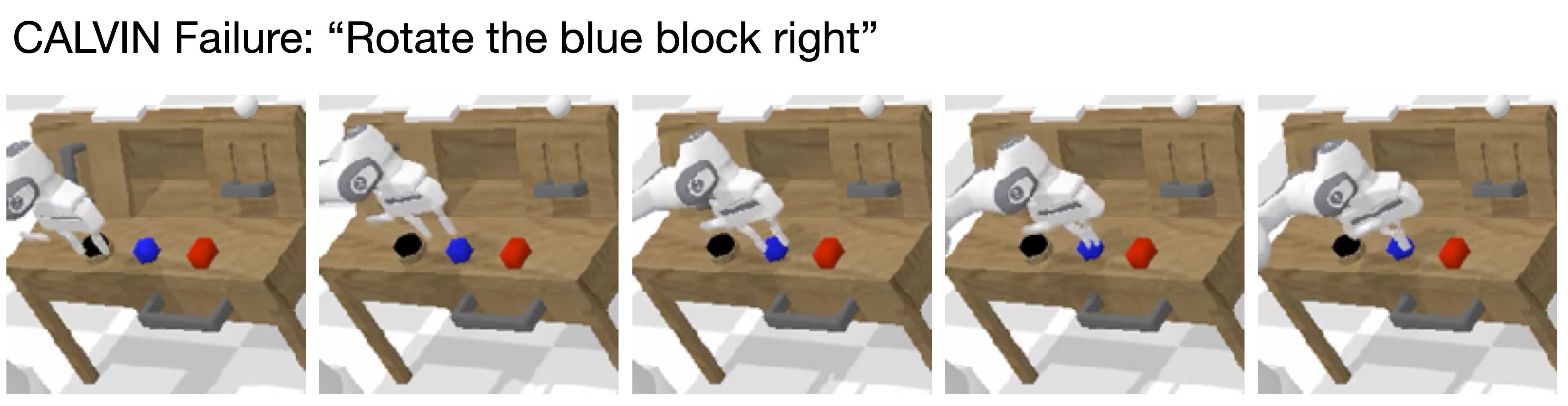}
        \caption{The robot attempts to grasp the blue block but grasps too high, failing to pick the block.}
    \end{subfigure}
    \begin{subfigure}[t]{0.49\textwidth}
        \includegraphics[width=\textwidth]{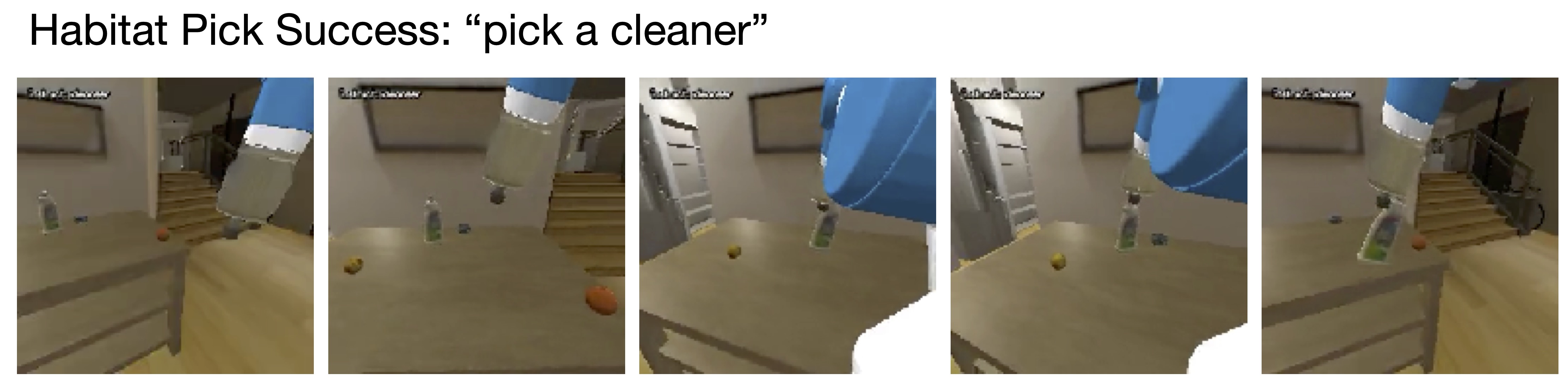}
        \caption{The robot moves closer to the cleaner bottle with its base and moves the arm to grasp the cleaner. It then returns the end-effector to the resting position to successfully end the task.}
    \end{subfigure}
    \begin{subfigure}[t]{0.49\textwidth}
        \includegraphics[width=\textwidth]{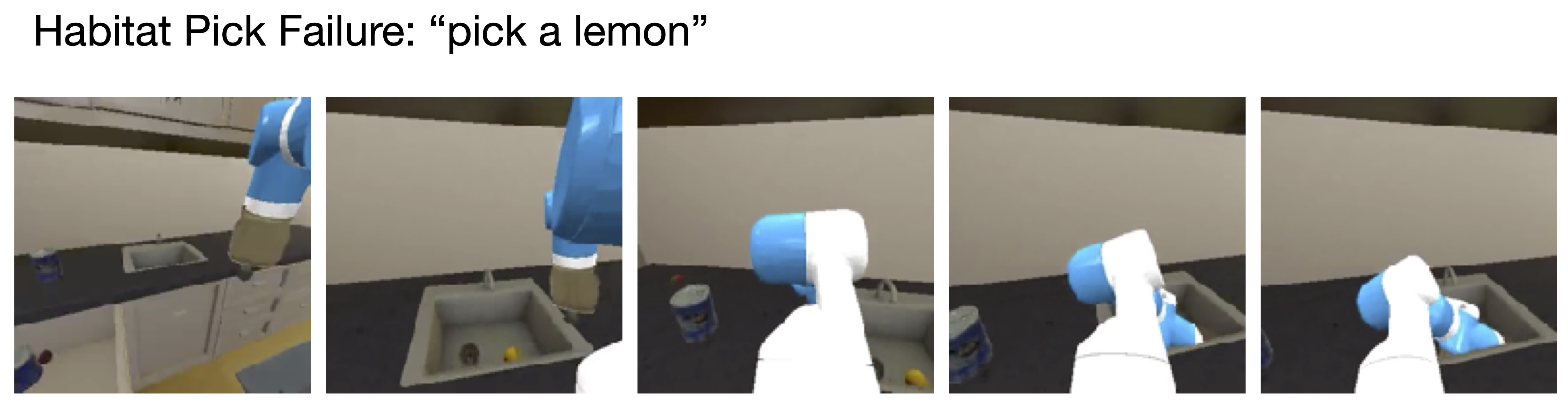}
        \caption{The robot correctly finds the lemon in the sink, but the tight sink receptacle results in the arm colliding with the sink and the episode terminating due to excessive collisions.}
    \end{subfigure}
    \begin{subfigure}[t]{0.49\textwidth}
        \includegraphics[width=\textwidth]{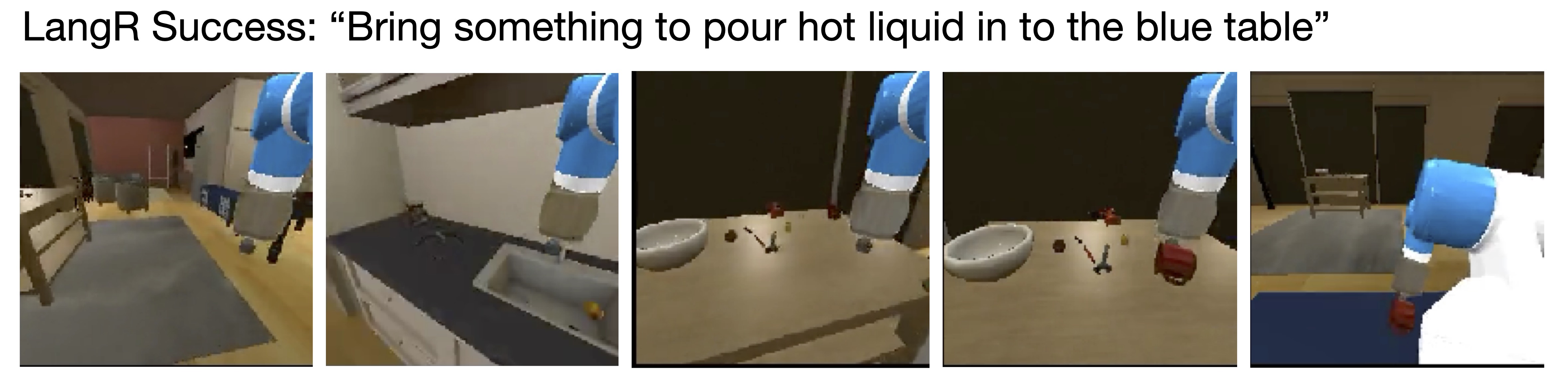}
        \caption{The robot searches the house, eventually finds the mug and then brings it to the blue table.}
    \end{subfigure}
    \begin{subfigure}[t]{0.49\textwidth}
        \includegraphics[width=\textwidth]{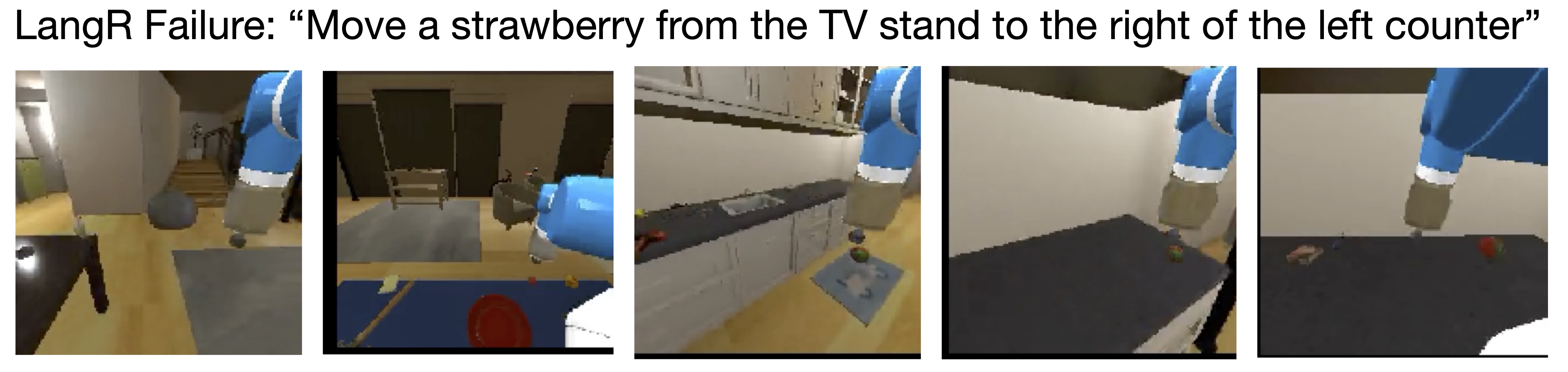}
        \caption{The robot picks the strawberry navigates to the counter area, but puts the strawberry on the right counter as opposed to the correct receptacle of the sink.}
    \end{subfigure}
    \caption{
        Qualitative visualizations of successes and failures from the results in Figure 1 of the main paper. The RVQ action space adapter is visualized for Meta-World, CALVIN and Habitat Pick. SemLang is visualized for Language Rearrangement.
    }
    \label{fig:qual-results} 
\end{figure*}

See \Cref{fig:qual-results} for qualitative results of results from \Cref{fig:vlm-disc}. The \rvq \asa is visualized for Meta-World, CALVIN and Habitat Pick. \semlang is visualized for Language Rearrangement.

\section{Per-Task Breakdown}
\label{sec:per-task} 

In this section, we show results for each environment by task type. \Cref{table:langr_full_results} shows performance on Language Rearrangement for each of the evaluation datasets. \Cref{table:calvin-full} shows performance on \calvin for each of the \calvin tasks. \Cref{table:babyai-full} shows performance on BabyAI for each of the BabyAI instruction types. \Cref{table:metaworld-full} shows performance on \metaworld for each of the 45 \metaworld task types.

\begin{table}[h!] 
  \resizebox{\columnwidth}{!}{
      \begin{tabular}{l|ccc|ccccccccccc|}
        \toprule
        & \multicolumn{3}{c|}{Aggregated} & \multicolumn{11}{c|}{Per Dataset Breakdown} \\
        & \textbf{Total} & \textbf{Behavior} & \textbf{Paraphrastic} & \textbf{Train} & \textbf{Scene} & \textbf{Instruct} & \textbf{Novel} & \textbf{Multiple} & \textbf{Referring} & \textbf{Context} & \textbf{Irrelevant} & \textbf{Multiple} & \textbf{Spatial} & \textbf{Conditional} \\
        & & \textbf{Generalization} & \textbf{Robustness} & & & \textbf{Rephrasing} & \textbf{Objects} & \textbf{Rearrange} & \textbf{Expressions} & & \textbf{Text} & \textbf{Objects} & & \textbf{Instructs} \\
        \midrule \midrule
        \input{tables/langR_full} 
      \end{tabular}
  }
  \caption{
    Evaluation results at 20M steps of RL training for all results in Language Rearrangement. We show averages and standard deviations over 2 random seeds of full policy training.
  }
  \label{table:langr_full_results} 
\end{table}

\begin{table}[h!] 
  \centering
    \input{tables/calvin_full}
  \caption{Breakdown on every \calvin task. Note there are not an equal proportion of all tasks in the evaluation dataset.}
  \label{table:calvin-full} 
\end{table}

\begin{table}[h!] 
  \centering
    \input{tables/gflan_full}
  \caption{Breakdown on every BabyAI task.}
  \label{table:babyai-full} 
\end{table}

\begin{table}[h!] 
  \centering
    \input{tables/metaworld_full}
  \caption{Breakdown on every \metaworld task.}
  \label{table:metaworld-full} 
\end{table}

%% file: tables/hyperparams.tex
\begin{tabular}{r|cccc}
                       Hyperparameter &\calvin    & \metaworld&     BabyAI &\habpick \\
                        \hline \hline
                                   LR &$ 3e^{-4}$ & $ 3e^{-4}$& $ 3e^{-4}$ &$ 3e^{-4}$ \\
                            Optimizer &AdamW      &    AdamW  &      AdamW &AdamW \\
                     Number of Epochs &3          &      3    &         20 &20 \\
                   Batch Size Per GPU &32         &     32    &          8 &32 \\
                       Context Length &12         &      3    &         32 &3 \\
                       Max Gradient Norm & 1 & 1 & 1 & 1 \\

\end{tabular}

%% file: tables/langR_full.tex
\textbf{\semlang} &  51 {\scriptsize $ \pm$ 1  } &  56 {\scriptsize $ \pm$ 2  } &  47 {\scriptsize $ \pm$ 1  } &  94 {\scriptsize $ \pm$ 3  } &  94 {\scriptsize $ \pm$ 6  } &  92 {\scriptsize $ \pm$ 1  } &  97 {\scriptsize $ \pm$ 0  } &  80 {\scriptsize $ \pm$ 6  } &  31 {\scriptsize $ \pm$ 3  } &  46 {\scriptsize $ \pm$ 14  } &  66 {\scriptsize $ \pm$ 6  } &  2 {\scriptsize $ \pm$ 2  } &  0 {\scriptsize $ \pm$ 0  } &  46 {\scriptsize $ \pm$ 4  } \\
\textbf{\lang} &  27 {\scriptsize $ \pm$ 12  } &  31 {\scriptsize $ \pm$ 14  } &  24 {\scriptsize $ \pm$ 10  } &  72 {\scriptsize $ \pm$ 13  } &  58 {\scriptsize $ \pm$ 11  } &  74 {\scriptsize $ \pm$ 12  } &  76 {\scriptsize $ \pm$ 29  } &  21 {\scriptsize $ \pm$ 10  } &  10 {\scriptsize $ \pm$ 12  } &  12 {\scriptsize $ \pm$ 11  } &  20 {\scriptsize $ \pm$ 13  } &  0 {\scriptsize $ \pm$ 0  } &  2 {\scriptsize $ \pm$ 3  } &  26 {\scriptsize $ \pm$ 16  } \\
\textbf{\mlp} &  42 {\scriptsize $ \pm$ 2  } &  45 {\scriptsize $ \pm$ 3  } &  38 {\scriptsize $ \pm$ 1  } &  99 {\scriptsize $ \pm$ 1  } &  96 {\scriptsize $ \pm$ 4  } &  92 {\scriptsize $ \pm$ 2  } &  95 {\scriptsize $ \pm$ 4  } &  47 {\scriptsize $ \pm$ 5  } &  26 {\scriptsize $ \pm$ 2  } &  34 {\scriptsize $ \pm$ 2  } &  32 {\scriptsize $ \pm$ 2  } &  0 {\scriptsize $ \pm$ 1  } &  8 {\scriptsize $ \pm$ 1  } &  39 {\scriptsize $ \pm$ 3  } \\
\midrule 

%% file: tables/calvin_full.tex
\begin{tabular}{ccccc}
\toprule
 & \textbf{\rvq} & \textbf{\mlp} & \textbf{\vq} & \textbf{\unif} \\
\midrule
\textbf{CALVIN} &  72  &  68  &  56  &  28  \\
\textbf{turn off led} &  50  &  96  &  36  &  16  \\
\textbf{move slider left} &  99  &  100  &  100  &  15  \\
\textbf{rotate red block right} &  54  &  17  &  35  &  17  \\
\textbf{open drawer} &  100  &  100  &  56  &  100  \\
\textbf{rotate red block left} &  31  &  14  &  14  &  14  \\
\textbf{push pink block right} &  31  &  100  &  51  &  14  \\
\textbf{push blue block right} &  42  &  27  &  35  &  20  \\
\textbf{push red block left} &  68  &  36  &  61  &  17  \\
\textbf{push pink block left} &  47  &  50  &  86  &  14  \\
\textbf{push red block right} &  35  &  35  &  17  &  17  \\
\textbf{push blue block left} &  56  &  27  &  47  &  14  \\
\textbf{push into drawer} &  49  &  34  &  14  &  14  \\
\textbf{rotate pink block left} &  76  &  73  &  73  &  16  \\
\textbf{turn on lightbulb} &  80  &  34  &  19  &  9  \\
\textbf{rotate pink block right} &  30  &  73  &  19  &  10  \\
\textbf{rotate blue block right} &  28  &  13  &  13  &  13  \\
\textbf{turn off lightbulb} &  76  &  19  &  19  &  12  \\
\textbf{lift blue block table} &  34  &  25  &  34  &  16  \\
\textbf{close drawer} &  100  &  100  &  100  &  70  \\
\textbf{rotate blue block left} &  32  &  11  &  38  &  20  \\
\textbf{move slider right} &  100  &  100  &  100  &  19  \\
\textbf{turn on led} &  31  &  100  &  42  &  14  \\
\textbf{lift blue block slider} &  32  &  22  &  51  &  15  \\
\textbf{lift pink block table} &  66  &  68  &  82  &  11  \\
\textbf{lift red block slider} &  56  &  22  &  41  &  13  \\
\textbf{lift red block table} &  45  &  53  &  15  &  15  \\
\textbf{lift pink block slider} &  75  &  12  &  62  &  12  \\

\end{tabular}

%% file: tables/gflan_full.tex
\begin{tabular}{cccc}
\toprule
 & \textbf{\semlang} & \textbf{\lang} & \textbf{\mlp} \\
\midrule
\textbf{goto} &  90  &  90  &  75  \\
\textbf{pickup} &  60  &  35  &  35  \\
\textbf{open} &  26  &  7  &  21  \\
\textbf{putnext} &  8  &  5  &  7  \\
\textbf{pick up seq go to} &  21  &  12  &  22  \\

\end{tabular}

%% file: tables/metaworld_full.tex
\begin{tabular}{ccccc}
\toprule
 & \textbf{\rvq} & \textbf{\mlp} & \textbf{\vq} & \textbf{\unif} \\
\midrule
\textbf{Meta-World} &  84  &  61  &  58  &  75  \\
\textbf{assembly} &  100  &  70  &  10  &  60  \\
\textbf{basketball} &  90  &  70  &  60  &  100  \\
\textbf{button-press-topdown} &  100  &  90  &  40  &  100  \\
\textbf{button-press-topdown-wall} &  100  &  100  &  60  &  90  \\
\textbf{button-press} &  100  &  100  &  70  &  100  \\
\textbf{button-press-wall} &  100  &  100  &  100  &  100  \\
\textbf{coffee-button} &  100  &  100  &  100  &  100  \\
\textbf{coffee-pull} &  100  &  40  &  30  &  50  \\
\textbf{coffee-push} &  80  &  20  &  30  &  80  \\
\textbf{dial-turn} &  100  &  50  &  40  &  100  \\
\textbf{disassemble} &  60  &  30  &  30  &  50  \\
\textbf{door-close} &  100  &  100  &  100  &  100  \\
\textbf{door-open} &  100  &  100  &  100  &  100  \\
\textbf{drawer-close} &  100  &  100  &  100  &  100  \\
\textbf{drawer-open} &  100  &  100  &  60  &  100  \\
\textbf{faucet-open} &  100  &  100  &  100  &  100  \\
\textbf{faucet-close} &  100  &  90  &  100  &  60  \\
\textbf{hammer} &  100  &  40  &  50  &  20  \\
\textbf{handle-press-side} &  100  &  100  &  100  &  100  \\
\textbf{handle-press} &  100  &  100  &  90  &  100  \\
\textbf{handle-pull-side} &  40  &  10  &  10  &  10  \\
\textbf{handle-pull} &  70  &  20  &  50  &  30  \\
\textbf{lever-pull} &  60  &  40  &  50  &  40  \\
\textbf{peg-insert-side} &  60  &  70  &  0  &  40  \\
\textbf{peg-unplug-side} &  50  &  30  &  100  &  90  \\
\textbf{pick-out-of-hole} &  50  &  90  &  40  &  30  \\
\textbf{pick-place} &  80  &  20  &  40  &  60  \\
\textbf{pick-place-wall} &  80  &  20  &  40  &  30  \\
\textbf{plate-slide} &  100  &  60  &  40  &  100  \\
\textbf{plate-slide-side} &  100  &  100  &  100  &  90  \\
\textbf{plate-slide-back} &  100  &  90  &  10  &  100  \\
\textbf{plate-slide-back-side} &  100  &  20  &  100  &  100  \\
\textbf{push-back} &  50  &  30  &  20  &  20  \\
\textbf{push} &  60  &  20  &  70  &  90  \\
\textbf{push-wall} &  80  &  40  &  60  &  100  \\
\textbf{reach} &  30  &  20  &  10  &  70  \\
\textbf{reach-wall} &  80  &  80  &  80  &  70  \\
\textbf{shelf-place} &  50  &  20  &  10  &  10  \\
\textbf{soccer} &  40  &  0  &  60  &  40  \\
\textbf{stick-push} &  100  &  60  &  10  &  100  \\
\textbf{stick-pull} &  90  &  50  &  30  &  100  \\
\textbf{sweep-into} &  70  &  40  &  60  &  70  \\
\textbf{sweep} &  90  &  30  &  50  &  70  \\
\textbf{window-open} &  100  &  100  &  100  &  100  \\
\textbf{window-close} &  100  &  100  &  100  &  100  \\

\end{tabular}

%% file: sections/checklist.tex
\clearpage
\section*{NeurIPS Paper Checklist}

\begin{enumerate}

\item {\bf Claims}
    \item[] Question: Do the main claims made in the abstract and introduction accurately reflect the paper's contributions and scope?
    \item[] Answer: \answerYes{} %
    \item[] Justification: Yes, our claims in the abstract and introduce are experimentally verified in \Cref{sec:experiments}.

\item {\bf Limitations}
    \item[] Question: Does the paper discuss the limitations of the work performed by the authors?
    \item[] Answer: \answerYes{} %
    \item[] Justification: Yes, we mention limitations in \Cref{sec:conclusion}.

\item {\bf Theory Assumptions and Proofs}
    \item[] Question: For each theoretical result, does the paper provide the full set of assumptions and a complete (and correct) proof?
    \item[] Answer: \answerNA{} %
    \item[] Justification: The paper does not include theoretical results.

    \item {\bf Experimental Result Reproducibility}
    \item[] Question: Does the paper fully disclose all the information needed to reproduce the main experimental results of the paper to the extent that it affects the main claims and/or conclusions of the paper (regardless of whether the code and data are provided or not)?
    \item[] Answer: \answerYes{} %
    \item[] Justification: Yes, we describe all experimental settings in detail in \Cref{sec:hyperparams}, we build on open-source models and benchmarks and we plan to open-source the code.

\item {\bf Open access to data and code}
    \item[] Question: Does the paper provide open access to the data and code, with sufficient instructions to faithfully reproduce the main experimental results, as described in supplemental material?
    \item[] Answer: \answerYes{} %
    \item[] Justification: We train and evaluate our method only on open benchmarks and train from an open source model. We also plan to release our code. 

\item {\bf Experimental Setting/Details}
    \item[] Question: Does the paper specify all the training and test details (e.g., data splits, hyperparameters, how they were chosen, type of optimizer, etc.) necessary to understand the results?
    \item[] Answer: \answerYes{} %
    \item[] Justification: Details provided in \Cref{sec:hyperparams}. 

\item {\bf Experiment Statistical Significance}
    \item[] Question: Does the paper report error bars suitably and correctly defined or other appropriate information about the statistical significance of the experiments?
    \item[] Answer: \answerYes{} %
    \item[] Justification: In our reinforcement learning results in \Cref{sec:disc-asa} we show results over 2 seeds and display the standard deviation between seeds as a shaded region in the plot. For other experiments we only report results on 1 seed since we are finetuning 7B parameter models, making it  computationally intensive to run multiple seeds.

\item {\bf Experiments Compute Resources}
    \item[] Question: For each experiment, does the paper provide sufficient information on the computer resources (type of compute workers, memory, time of execution) needed to reproduce the experiments?
    \item[] Answer: \answerYes{} %
    \item[] Justification: Yes, complete details are proivded in \Cref{sec:training-details}.
    
\item {\bf Code Of Ethics}
    \item[] Question: Does the research conducted in the paper conform, in every respect, with the NeurIPS Code of Ethics \url{https://neurips.cc/public/EthicsGuidelines}?
    \item[] Answer: \answerYes{} %
    \item[] Justification: Yes, the paper adheres to the ethics code.

\item {\bf Broader Impacts}
    \item[] Question: Does the paper discuss both potential positive societal impacts and negative societal impacts of the work performed?
    \item[] Answer: \answerNA{} %
    \item[] Justification: The paper is on foundational research not tied to any particular application and we do not feel it is important to highlight any negative societal impacts of our work.
    
\item {\bf Safeguards}
    \item[] Question: Does the paper describe safeguards that have been put in place for responsible release of data or models that have a high risk for misuse (e.g., pretrained language models, image generators, or scraped datasets)?
    \item[] Answer: \answerNA{} %
    \item[] Justification: The paper poses no such risks.

\item {\bf Licenses for existing assets}
    \item[] Question: Are the creators or original owners of assets (e.g., code, data, models), used in the paper, properly credited and are the license and terms of use explicitly mentioned and properly respected?
    \item[] Answer: \answerYes{} %
    \item[] Justification: We cite all datasets used.

\item {\bf New Assets}
    \item[] Question: Are new assets introduced in the paper well documented and is the documentation provided alongside the assets?
    \item[] Answer: \answerNA{} %
    \item[] Justification: The paper does not release new assets.

\item {\bf Crowdsourcing and Research with Human Subjects}
    \item[] Question: For crowdsourcing experiments and research with human subjects, does the paper include the full text of instructions given to participants and screenshots, if applicable, as well as details about compensation (if any)? 
    \item[] Answer: \answerNA{} %
    \item[] Justification: the paper does not involve crowdsourcing nor research with human subjects.

\item {\bf Institutional Review Board (IRB) Approvals or Equivalent for Research with Human Subjects}
    \item[] Question: Does the paper describe potential risks incurred by study participants, whether such risks were disclosed to the subjects, and whether Institutional Review Board (IRB) approvals (or an equivalent approval/review based on the requirements of your country or institution) were obtained?
    \item[] Answer: \answerNA{} %
    \item[] Justification: The paper does not involve crowdsourcing nor research with human subjects.

\end{enumerate}